\documentclass{article}
\usepackage{ijcai24}

\usepackage{times}
\usepackage{soul}
\usepackage{url}
\usepackage[hidelinks]{hyperref}
\usepackage[utf8]{inputenc}
\usepackage[small]{caption}
\usepackage{graphicx}
\usepackage{amsmath}
\usepackage{amsthm}
\usepackage{booktabs}
\usepackage{algorithm}
\usepackage{algorithmic}
\usepackage[switch]{lineno}

\usepackage{amsmath,amssymb, dsfont, xcolor}
\usepackage{graphicx, subcaption}

\title{Cross-Domain Feature Augmentation for Domain Generalization}

\author{
Yingnan Liu$^{1,2}$
\and
Yingtian Zou$^{1,2}$\and
Rui Qiao$^1$\and
Fusheng Liu$^2$\and
Mong Li Lee$^{1,2}$ \And
Wynne Hsu$^{1,2}$\\
\affiliations
$^1$School of Computing, National University of Singapore\\
$^2$Institute of Data Science, National University of Singapore\\
\emails
\{liu.yingnan, yingtian, rui.qiao, fusheng\}@u.nus.edu,
\{leeml, whsu\}@comp.nus.edu.sg
}

\begin{document}

\maketitle

\maketitle

\begin{abstract} 
   Domain generalization aims to develop models that are robust to distribution shifts. 
   Existing methods focus on learning invariance across domains to enhance model robustness, and data augmentation has been widely used to learn invariant predictors, with most methods performing augmentation in the input space. 
   However, augmentation in the input space has limited diversity whereas in the feature space is more versatile and has shown promising results. 
   Nonetheless, feature semantics is seldom considered and existing feature augmentation methods suffer from a limited variety of augmented features. We decompose features into class-generic, class-specific, domain-generic, and domain-specific components.
   We propose a cross-domain feature augmentation method named XDomainMix that enables us to increase sample diversity while emphasizing the learning of invariant representations to achieve domain generalization.
   Experiments on widely used benchmark datasets demonstrate that our proposed method is able to achieve state-of-the-art performance. Quantitative analysis indicates that our feature augmentation approach facilitates the learning of effective models that are invariant across different domains. Our code is available at \url{https://github.com/NancyQuris/XDomainMix}.
   
\end{abstract}

\section{Introduction}  
\label{ch:in} 
Deep learning methods typically assume that training and testing data are independent and identically distributed. However, this assumption is often violated in the real world, leading to a decrease in model performance when faced with a different distribution \cite{torralba2011unbiased}. 
The field of domain generalization aims to mitigate this issue by learning a model from one or more distinct yet related training domains, with the goal of generalizing effectively to domains that have not been previously encountered.
Studies  suggest that the poor generalization on unseen distributions can be attributed to the failure of learning the \textit{invariance} across different domains during the training phase \cite{muandet2013domain,li2018domain}. To tackle this,  research has focused on representation learning and data augmentation as key to learning invariance.

Invariant representation learning aims to align representation across domains \cite{shi2022gradient}, and learn invariant causal predictors \cite{arjovsky2019invariant}. 
They usually impose regularization, which may result in a hard optimization problem \cite{pmlr-v162-yao22b}.
In contrast,  data augmentation techniques propose to generate additional samples for the learning of invariance, and avoid the complexities in the regularization approach \cite{mancini2020towards}.
Data augmentation can be generally classified into two types: input space and feature space augmentation. The former often encounters limitations due to a lack of diversity in the augmented data \cite{li2021simple}, while the latter offers more versatility and has yielded promising outcomes \cite{zhou2020domain}. 

\begin{figure}[t!]
        \centering
        \begin{subfigure}{0.3\linewidth}
            \centering
            \includegraphics[width=0.8\linewidth]{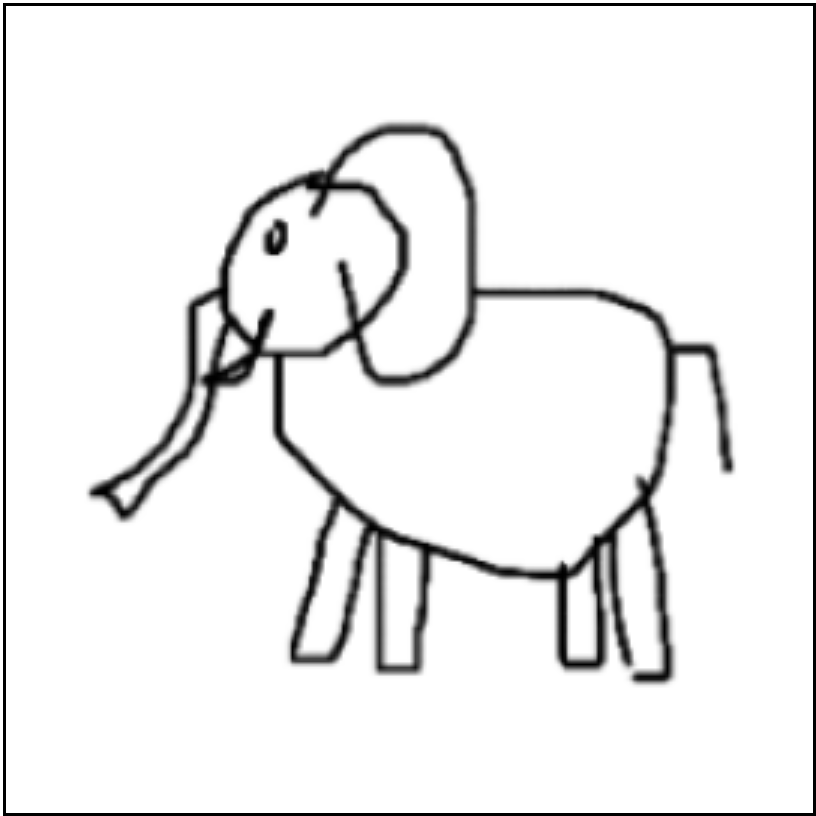}\\
            \includegraphics[width=0.8\linewidth]{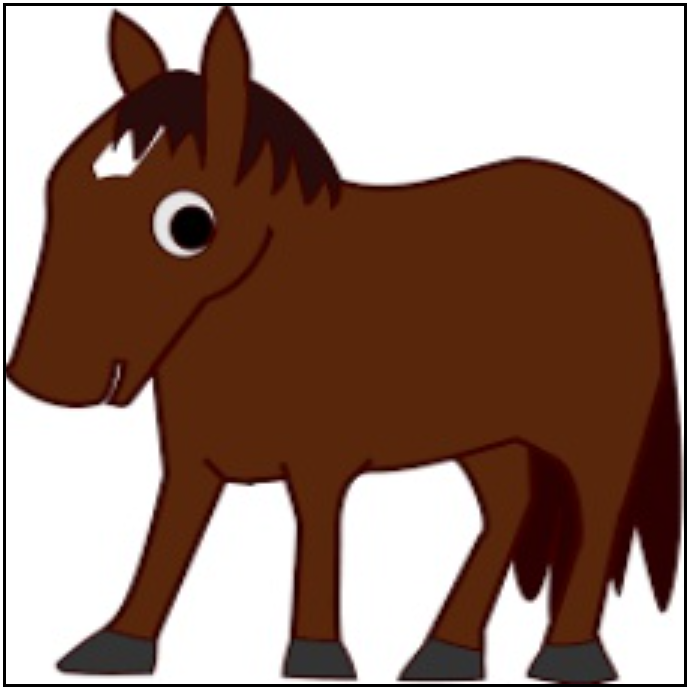}\\
            \includegraphics[width=0.8\linewidth]{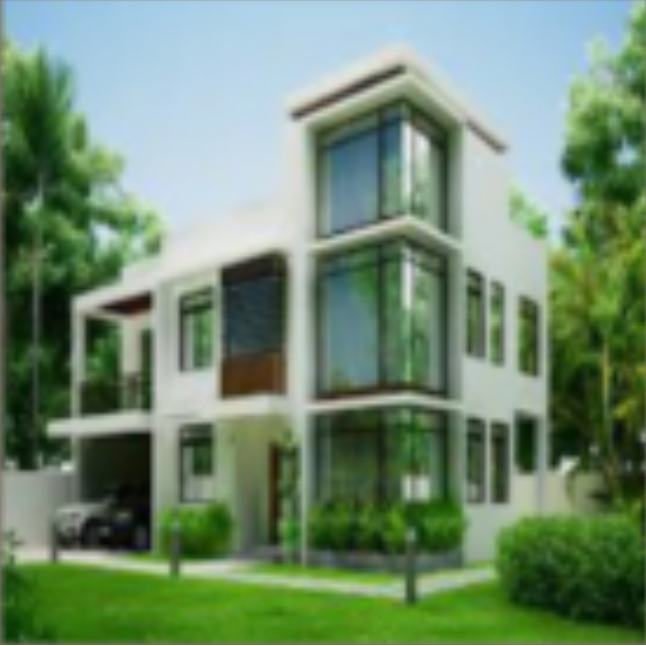}\\ 
            \caption{Original} 
        \end{subfigure}
        \begin{subfigure}{0.3\linewidth}
            \centering        
            \includegraphics[width=0.8\linewidth]{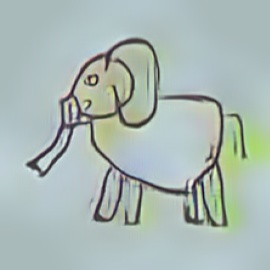}\\
            \includegraphics[width=0.8\linewidth]{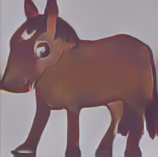}\\
            \includegraphics[width=0.8\linewidth]{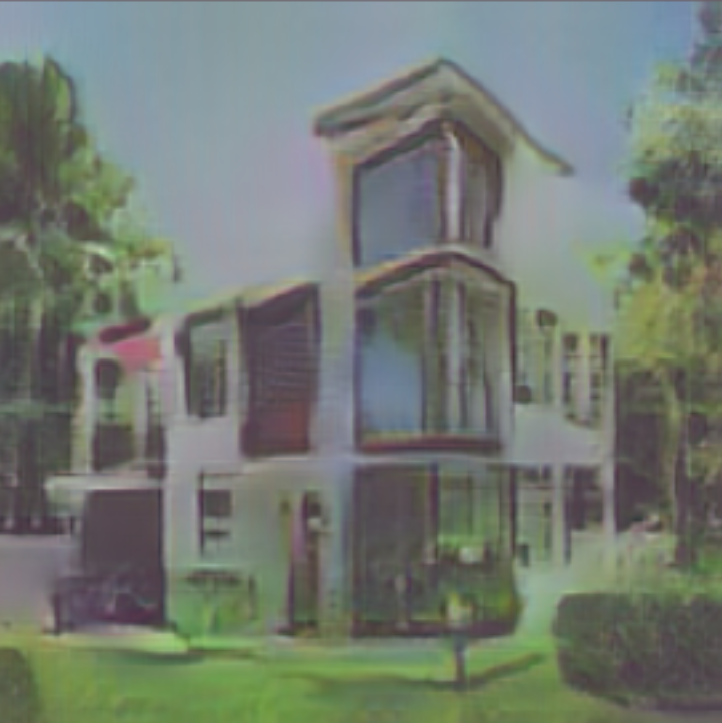}\\
            \caption{DSU} 
        \end{subfigure}
        \begin{subfigure}{0.3\linewidth}
            \centering
            \includegraphics[width=0.8\linewidth]{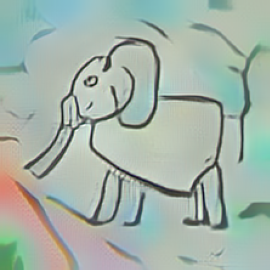}\\
            \includegraphics[width=0.8\linewidth]{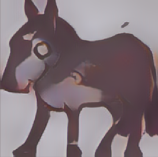}\\
            \includegraphics[width=0.8\linewidth]{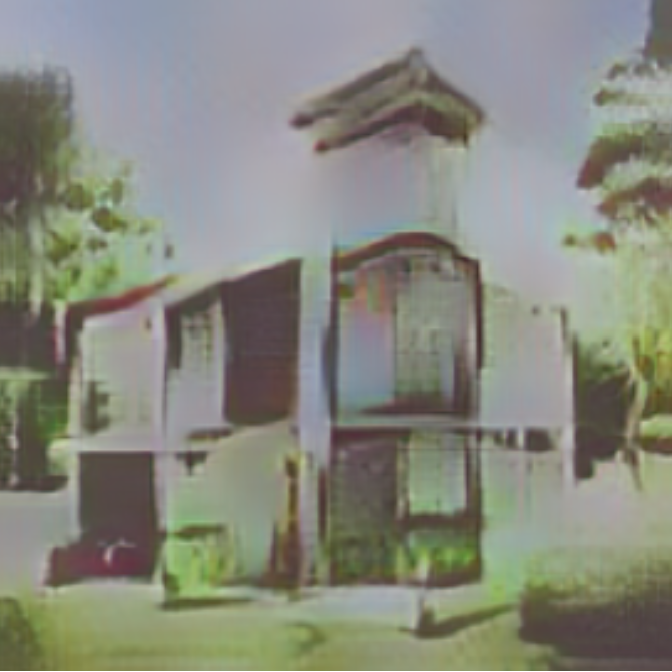}\\ 
            \caption{XDomainMix}
        \end{subfigure}
            
        \caption{Samples of images reconstructed from features produced by DSU ~\protect\cite{li2022uncertainty} and the proposed XDomainMix.
        The elephant reconstructed from XDomainMix's features shows a more complex background. The horse reconstructed from XDomainMix's features displays different characteristics such as a white belly. The top floor of the house generated by XDomainMix's features shows solid walls, instead of glass walls in the original. In contrast, images reconstructed from DSU's features have limited diversity and appear largely similar to the original images. } \label{fig:1}
    \end{figure}

Despite the versatility of feature space augmentation, existing methods such as MixStyle \cite{zhou2020domain} and DSU \cite{li2022uncertainty} do not consider feature semantics during the augmentation process. Instead, they alter feature statistics which  often leads to a limited range of diversity.
This lack of diversity in the generated features motivates us to decompose the features according to feature semantics.
We build on 
prior research which suggests that the features learned for each class can be viewed as a combination of class-specific and class-generic components \cite{chu2020feature}. The class-specific component carries information unique to a class, while the class-generic component carries information that is shared across classes.
We observe that, even within the same class, features of samples from different domains can be distinguished, indicating that these features may contain domain-specific information. As such, we broaden our understanding of features to include domain-specific and domain-generic components.

We introduce a method called XDomainMix that changes domain-specific components of a feature while preserving class-specific components. With this, the model is able to learn features that are not tied to specific domains, allowing it to make predictions based on features that are invariant across domains.
Figure \ref{fig:1} shows samples of original  images and reconstructed images based on  existing feature augmentation technique (DSU) and the proposed XDomainMix. We visualize the augmented features using a pre-trained autoencoder \cite{Huang_2017_ICCV}. 
From the reconstructed images, we see that DSU's augmented features remain largely the same as that of the original image feature.
On the other hand, the images reconstructed from the features obtained using XDomainMix have richer variety while preserving the salient features of the class. 

Results of experiments on benchmark datasets show the superiority of XDomainMix for domain generalization.
We quantitatively measure the invariance of learned representation and prediction to show that the models trained with XDomainMix's features are more invariant across domains compared to state-of-the-art feature augmentation methods.
Our measurement of the divergence between original features and augmented features shows that XDomainMix results in more diverse augmentation.

\section{Related Work}
To learn invariance, existing domain generalization approaches can be categorized into representation learning methods and data augmentation methods. 
Works on learning invariant representation employ regularizers to align representations or gradients \cite{sun2016deep,li2020domain_2,kim2021selfreg,mahajan2021domain,shi2022gradient,rame2022fishr,yao2022pcl} across different domains, 
enforce the optimal classifier on top of the representation space to be the same across all domains \cite{arjovsky2019invariant,ahuja2021invariance}, 
or uses distributionally robust optimization \cite{Sagawa*2020Distributionally}. 
However, the use of regularization terms during learning of invariant representation could
make the learning process more complex and potentially limit the model's expressive power.

Another approach is to employ data augmentation to learn invariance. 
Existing work that operates in the input space includes
network-learned transformation \cite{zhou2020learning,li2021progressive}, adversarial data augmentation \cite{volpi2018generalizing,shankar2018generalizing}, mixup \cite{mancini2020towards,pmlr-v162-yao22b}, and Fourier-based transformation \cite{xu2021fourier}.
Each of these techniques manipulates the input data in different ways to create variations that help the model learn invariant features.
However, the range of transformations that can be applied in the input space is often limited.

On the other hand, feature augmentation can offer more flexibility and potential for learning more effective invariant representations.
Prior work has generated diverse distributions in the feature space by changing feature statistics \cite{zhou2020domain,jeon2021feature,wang2022feature,li2022uncertainty,fan2023towards},
adding noise \cite{li2021simple}, or mixing up features from different domains \cite{mancini2020towards,qiao2021uncertainty}. For example,
MixStyle \cite{zhou2020domain} synthesizes new domains by mixing the feature statistics of two features. 
DSU \cite{li2022uncertainty} extends the idea by modeling feature statistics as a probability distribution and using new feature statistics drawn from the distribution to augment features.
In addition to generating diverse distributions, RSC \cite{huang2020self} adopts a different approach by discarding the most activated features instead of generating diverse data. This encourages the model to use less-activated features that might be associated with labels relevant to data outside the domain.

In contrast to existing methods, our work carefully considers feature semantics by leveraging class-label information and domain information to augment features. This increases intra-class variability and helps the model to learn a broader understanding of each class, thus improving its ability to handle new, unseen data.

\section{Proposed Method}
\begin{figure*}
    \centering
    \includegraphics[width=0.9\linewidth]{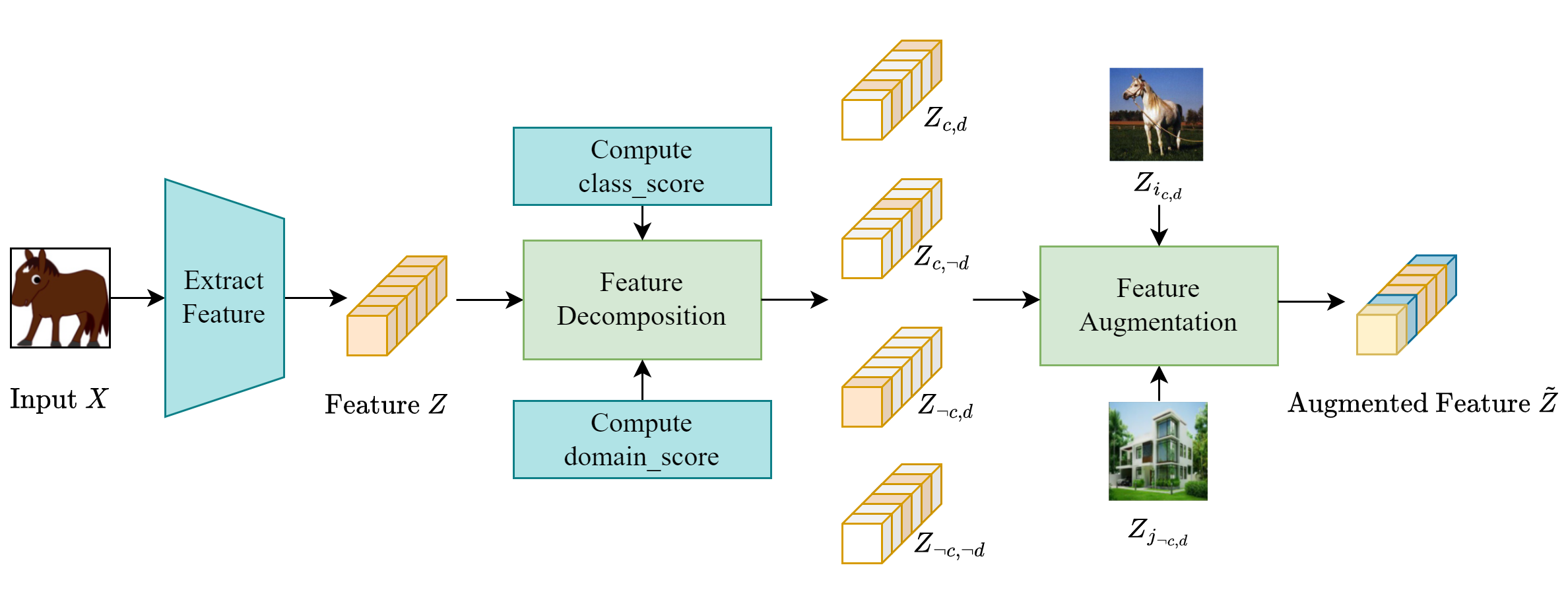}
    \caption{Overview of XDomainMix. To perform augmentation, the feature of an input is decomposed into four components based on the semantics' correlation with class and domain. Afterward, features of other two samples from different domains, one from the same class and one from a different class are used to augment features by changing domain-specific feature components. 
    }
    \label{fig:overview}
\end{figure*}
We consider the problem where we have a set of source domains $\mathcal{D}_S=\{S_1,\cdots,S_N\}$, $N>1$ and a set of unseen domains $\mathcal{D}_U$. Each source domain $S_i=\{(x_j^{(i)}, y_j^{(i)})\}_{j=1}^{n_i}$ has a joint distribution on the input $x$ and the label $y$. 
Domains in $\mathcal{D}_U$ have distinct joint distributions from those of the domains in $\mathcal{D}_S$.
We assume that all domains in $D_S$ and $D_U$ share the same label space but the class distribution across domains need not be the same. 
The goal is to learn a mapping $g:x\to y$ using the source domains in $\mathcal{D}_S$ such that the error is minimized when $g$  is applied to samples in $\mathcal{D}_U$.

In deep learning, $g$ is typically realized as a composition of two functions: a feature extractor $f:x\to Z$ that maps input $x$ to $Z$ in the latent feature space, followed by a classifier $c: Z \to y$ that maps $Z$ to the output label $y$. 
Ideally, $f$ should extract features that are domain invariant yet retain class-specific information. The features of a given input can be decomposed 
into two components: class-specific and class-generic. The class-specific component consists of feature semantics that are strongly correlated with class labels, making them more important in discriminating a target class from other classes.

Furthermore, features can also be decomposed into domain-specific and domain-generic components. This is because samples from different domains, even if they belong to the same class, possess unique feature characteristics to their respective domains.
We extend these notions to decompose the features extracted by $f$ into four distinct components:  class-specific domain-specific ($Z_{c,d}$), class-specific domain-generic ($Z_{c,\neg d}$), class-generic domain-specific ($Z_{\neg c,d}$), and class-generic domain-generic ($Z_{\neg c,\neg d}$) component. 

We determine whether an extracted feature contains information that is specific to a class or a domain by its importance to the prediction of the class and domain respectively.
In other words, if a feature is important to the prediction of a specific class and a specific domain, it is considered a class-specific domain-specific component. If a feature is crucial in class prediction but not domain prediction, it falls into the class-specific domain-generic category. Similarly, features that are important for domain but not class predictions are categorized as class-generic domain-specific, and those not significant to either are labeled as class-generic domain-generic.

Our proposed feature augmentation strategy, XDomainMix, performs augmentation in the feature space by modifying the domain-specific component of features in a way that preserves class-related information. To discourage the use of domain-specific features for class prediction and encourage the exploitation of less activated features, class-specific domain-specific component is discarded with some probability during training. Details of feature decomposition and augmentation are described in the following subsections.
Figure \ref{fig:overview} shows an overview of our proposed method.

\subsection{Feature Decomposition}
Suppose the feature extractor $f$ extracts $Z=f(x)\in \mathbb{R}^K$. Let $z^k$ be the $k^{th}$ dimension of $Z$. 
We determine whether $z^k$ is class-specific or class-generic via the class importance score, which is computed as the product of feature value and the derivative of class classifier $c$'s predicted logit $v_c$ of $x$'s ground truth class \cite{selvaraju2017grad,chu2020feature} as they show how much $z^k$ contributes to $v_c$: 
\begin{equation}
    \text{class\_score}(z^k) = \frac{\partial v_c}{\partial z^k}z^k
    \label{eq:class_im}
\end{equation}
To determine if $z^k$ is domain-specific, we employ a domain classifier $d$ that has an identical architecture as the class classifier $c$. 
$d$ is trained to predict domain labels of features extracted by the feature extractor.
Similar to Equation \ref{eq:class_im}, domain importance score is computed using the derivative of $d$'s predicted logit $v_d$ of $x$'s ground truth domain:
\begin{equation}
    \text{domain\_score}(z^k) = \frac{\partial v_d}{\partial z^k}z^k
    \label{eq:domain_im}
\end{equation}

Let $\tau_c$ and $\tau_d$ be predefined thresholds 
for filtering feature dimensions that are class-specific and domain-specific respectively. We obtain a class-specific mask $M_c\in \mathbb{R}^K$ and a domain-specific mask $M_d\in \mathbb{R}^K$ on $Z$ for $\{z^k\}_{k=1}^K$ where their respective $k^{th}$ entries are given as follows: 
\begin{equation}
    \begin{split}
    M_c[k] &= 
    \begin{cases}
        1 & \text{if }  \text{class\_score}(z^k) > \tau_c\\
        0 & \text{otherwise} 
    \end{cases},  \\
    M_d[k] &= 
    \begin{cases}
        1 & \text{if }  \text{domain\_score}(z^k) > \tau_d\\
        0 & \text{otherwise} 
    \end{cases}
    \end{split}
    \label{eq:mask}
\end{equation}

Complementary class-generic mask and domain-generic mask are obtained by $\mathds{1}-M_c$ and $\mathds{1}-M_d$ where $\mathds{1}$ is the tensor of values 1 and of the same size as $Z$. 
Class-specific domain-specific ($Z_{c,d}$), class-specific domain-generic ($Z_{c,\neg d}$), class-generic domain-specific ($Z_{\neg c,d}$), and class-generic domain-generic ($Z_{\neg c,\neg d}$) feature components are obtained by 
\begin{equation}
    \begin{split}
        Z_{c,d} &= M_c \odot M_d \odot Z \\
        Z_{c,\neg d} &= M_c \odot (\mathds{1}-M_d) \odot Z \\
        Z_{\neg c,d} &= (\mathds{1}-M_c) \odot M_d \odot Z \\
        Z_{\neg c,\neg d} &= (\mathds{1}-M_c) \odot (\mathds{1}-M_d) \odot Z
    \end{split}
    \label{eq:decom}
\end{equation}
where $\odot$ is element-wise multiplication. Note that $Z_{c,d} + Z_{c,\neg d} + Z_{\neg c,d} + Z_{\neg c,\neg d} = Z$.

\subsection{Cross Domain Feature Augmentation}
 To achieve domain invariance and reduce reliance on domain-specific information presented in training domains during prediction, we manipulate domain-specific feature components to enhance diversity from a domain perspective. 
 Further, the augmentation should increase feature diversity while preserving class semantics using existing data. This is achieved by mixing the class-specific domain-specific feature component of a sample  with the class-specific domain-specific feature component of a same-class sample from other domains. 
For class-generic domain-specific feature component, it is mixed with the class-generic domain-specific feature component of a different-class sample from other domains, introducing further  diversity.

Specifically, for the feature $Z$ extracted from input $x$, we randomly sample two inputs $x_i$ and $x_j$ whose domains are different from $x$. Further,  $x_i$ has the same class label as $x$ while $x_j$ is from a different class. 
Let $Z_i$ be the feature extracted from input $x_i$ and $Z_j$ be the feature extracted from  $x_j$. 
Then we have
\begin{equation}
    \begin{split}
        \tilde{Z}_{c,d} &= \lambda_1 Z_{c,d} + (1-\lambda_1) {Z_i}_{c,d} , \\
        \tilde{Z}_{\neg c,d} &= \lambda_2 Z_{\neg c,d} + (1-\lambda_2) {Z_j}_{\neg c,d}
    \end{split}
    \label{eq:mix}
\end{equation}
where $\lambda_1$ and $ \lambda_2$ are the mixup ratios independently sampled from a uniform distribution $U(0, 1)$. 

A new feature $\tilde{Z}$ with the same class label as $Z$ is generated by replacing the domain-specific component in $Z$ as follows:
\begin{equation}
    \tilde{Z}= 
        \tilde{Z}_{c,d} + \tilde{Z}_{\neg c,d} + Z_{c,\neg d} + Z_{\neg c,\neg d}
\end{equation}

To further encourage the model to focus on domain-invariant features and exploit the less activated feature during class prediction, we discard the class-specific domain-specific feature component with some probability $p_\text{discard}$ as follows: 
\begin{equation}
    \tilde{Z}= 
    \begin{cases}
        \tilde{Z}_{\neg c,d} + Z_{c,\neg d} + Z_{\neg c,\neg d} & \text{if }  p \leq p_\text{discard}\\
        \tilde{Z}_{c,d} + \tilde{Z}_{\neg c,d} + Z_{c,\neg d} + Z_{\neg c,\neg d} & \text{otherwise}
    \end{cases}
    \label{eq:discard}
\end{equation}
where $p$ is randomly sampled from a uniform distribution $U(0, 1)$.

\subsection{Training Procedure}
Prior research has shown that empirical risk minimization (ERM) \cite{vapnik1999nature} is a competitive baseline \cite{gulrajani2021in,wiles2022a}. The objective function of ERM is given by
\begin{equation}
    \mathcal{L}_{erm}=\frac{1}{N}\sum_{i=1}^{N}\frac{1}{n_i}\sum_{j=1}^{n_i}\ell(\hat{y}_j^{(i)}, y_j^{(i)})
\end{equation}
where $\ell$ is the loss function to measure the error between the predicted class $\hat{y}_j^{(i)}$ and the ground truth $y_j^{(i)}$. $N$ is the number of training domains and $n_i$ is the number of training samples in domain $i$. 

We train the model in two phases.
During the warm-up phase, the feature extractor $f$ and class classifier $c$ are trained on the original dataset for class label prediction following $\mathcal{L}_{erm}$. The domain classifier $d$ is trained using ${Z}$, the features extracted by $f$, to predict domain labels. 
 The objective function of $d$ is given by 
\begin{equation}
    \mathcal{L}_d = \frac{1}{N}\sum_{i=1}^{N}\frac{1}{n_i}\sum_{j=1}^{n_i}\ell(d({Z}_j^{(i)}),i)
\end{equation}
where $\ell$ is the loss function to measure the error between the predicted domain $d({Z}_j^{(i)})$ and the ground truth $i$. 

When the warm-up phase is completed, we use Equation \ref{eq:decom} to decompose the features obtained
from $f$. Augmented features are then derived using Equation \ref{eq:discard}.
Both feature extractor $f$ and class classifier $c$ are trained using the original and augmented features with the following objective function:
\begin{equation}
\begin{split}
    \mathcal{L}_{aug} = \frac{1}{N} \sum_{i=1}^{N}
    \frac{1}{2n_i} \sum_{j=1}^{n_i}
    \left[ \ell(c({Z}_j^{(i)}), y_j^{(i)}) + 
    \ell(c(\tilde{Z}_j^{(i)}), y_j^{(i)}) \right]
\end{split}        
\end{equation}
where  $\tilde{Z}_j^{(i)}$ is the augmented feature derived from $Z_j^{(i)}$.  $c({Z}_j^{(i)})$ is the predicted class given ${Z}_j^{(i)}$, and $c(\tilde{Z}_j^{(i)})$ is the predicted class given $\tilde{Z}_j^{(i)}$. $y_j^{(i)}$ is the ground truth class. 

We also train the domain classifier $d$ using $\mathcal{L}_d$.
Note that the domain classifier $d$ is not trained using this set of augmented features, as the augmented features do not have assigned domain labels since they need not follow the distribution of existing domains.

\section{Performance Study}

 We implement our proposed solution using PyTorch 1.12.0 and perform a series of experiments on NVIDIA Tesla V100 GPU to evaluate the effectiveness of the proposed XDomainMix. The following benchmark datasets are used:
\begin{itemize}
    \item Camelyon17 \cite{bandi2018detection} from Wilds \cite{WILDS}. This dataset contains 455,954 tumor and normal tissue slide images from 5 hospitals (domains). Distribution shift arises from variations in patient population, slide staining, and image acquisition. 

     \item FMoW \cite{christie2018functional} from Wilds. This dataset contains 141,696 satellite images from 62 land use categories across 16 years from 5 regions (domains). 
    
    \item PACS \cite{li2017deeper}. This dataset contains 9,991 images of 7 objects in 4 visual styles (domains): art painting, cartoon, photo, and sketch. 

    \item TerraIncognita \cite{beery2018recognition}. The dataset contains 24,788 images from 10 categories of wild animals taken from 4 different locations (domains). 

    \item DomainNet \cite{peng2019moment}. This dataset contains 586,575 images from 365 classes in 6 visual styles (domains): clipart, infograph, painting, quickdraw, real, and sketch.

\end{itemize}

\begin{table*}[t!]
    \centering
  
    \begin{tabular}{l|lllll}
        \hline
        Method  & Camelyon17      & FMoW            & PACS                    & TerraIncognita  & DomainNet     \\ \hline
        ERM     & 70.3$\pm$6.4    & 32.3$\pm$1.3    & 85.5$\pm$0.2            & 46.1$\pm$1.8    & 43.8$\pm$0.1          \\
        GroupDRO& 68.4$\pm$7.3    & 30.8$\pm$0.8    & 84.4$\pm$0.8           & 43.2$\pm$1.1    & 33.3$\pm$0.2          \\
        RSC     & 77.0$\pm$4.9\^  & 32.6$\pm$0.5\^  & 85.2$\pm$0.9             & 46.6$\pm$1.0    & 38.9$\pm$0.5          \\
        MixStyle& 62.6$\pm$6.3\^  & 32.9$\pm$0.5\^  & 85.2$\pm$0.3             & 44.0$\pm$0.7    & 34.0$\pm$0.1          \\
        DSU     & 69.6$\pm$6.3\^  & 32.5$\pm$0.6\^  & 85.5$\pm$0.6\^       & 41.5$\pm$0.9\^  & 42.6$\pm$0.2\^{}      \\
        LISA    & 77.1$\pm$6.5    & 35.5$\pm$0.7    & 83.1$\pm$0.2\^        & 47.2$\pm$1.1\^  & 42.3$\pm$0.3\^{}        \\
        Fish    & 74.7$\pm$7.1    & 34.6$\pm$0.2    & 85.5$\pm$0.3             & 45.1$\pm$1.3    & 42.7$\pm$0.2          \\ 
        \textbf{XDomainMix} & \textbf{80.9$\pm$3.2} & \textbf{35.9$\pm$0.8} & \textbf{86.4$\pm$0.4}  &  \textbf{48.2$\pm$1.3}    & \textbf{44.0$\pm$0.2}      \\ \hline
    \end{tabular}
   \caption{Domain generalization performance of XDomainMix compared with state-of-the-art methods performance published in ~\protect\cite{pmlr-v162-yao22b,gulrajani2021in,cha2021swad}.  Results with \^{} are produced by us. }
    \label{tb:resultA}
\end{table*} 

The class importance thresholds $\tau_c$ and domain importance thresholds $\tau_d$ in Equation \ref{eq:mask} are set as follows:
$\tau_c$ is set to be the 50\%-quantile of the class importance scores of $\{z^k\}$ in a feature $Z$ so that 50\% dimensions are considered class-specific, while the remaining 50\% are class-generic.
$\tau_d$ controls the strength of the augmentation as it determines the identification of domain-specific feature components. 
We employ a cyclic changing scheme for $\tau_d$ to let the model learn gradually from weak augmentation to strong augmentation and give the domain classifier more time to adapt to a more domain-invariant feature extractor. 
The value is initially set to be 90\%-quantile of domain importance scores of $\{z^k\}$ in a feature $Z$. As the training proceeds, $\tau_d$ is decreased by 10\% quantile for every $n$ step until it reaches the 50\%-quantile, where it also remains for $n$ steps. The same cycle is repeated where $\tau_d$ is set to be 90\%-quantile again. 
Input $x_i, x_j$ used in augmentation (Equation \ref{eq:mix}) are samples from the same training batch. $p_\text{discard}$ is set to 0.2.

For Camelyon17 and FMoW datasets, we follow the setup in LISA \cite{pmlr-v162-yao22b}. Non-pretrained DenseNet-121 is used for Canmelyon17 and pretrained DenseNet-121 is used for FMoW.
We use the same partitioning in  Wilds \cite{WILDS} to obtain the training, validation, and test domains.
The batch size is set to 32, and the model is trained for 2 epochs for Camelyon17 and 5 epochs for FMoW. The learning rate and weight decay are set to 1e-4 and 0. The warm-up phase is set to 4000 steps. We tune the step $n$ in $\{100, 500\}$ for changing $\tau_d$. The best model is selected based on its performance in the validation domain.

For PACS, 
TerraIncognita and DomainNet datasets, we follow the setup in DomainBed \cite{gulrajani2021in}, and use a pre-trained ResNet-50. Each domain in the dataset is used as a test domain in turn, with the remaining domains serving as training domains.  The batch size is set to 32 (24 for DomainNet), and the model is trained for 5000 steps (15000 steps for DomainNet). We tune the learning rate in \{2e-5, 3e-5, 4e-5, 5e-5, 6e-5\} and weight decay in (1e-6, 1e-2) using the DomainBed framework. The warm-up phase is set to 3000 steps and $n$ is set to 100 steps. The best model is selected based on its performance on the validations split of the training domains.


\subsection{Domain Generalization Performance}
We compare our proposed XDomainMix with ERM \cite{vapnik1999nature} and the following state-of-the-art methods: 
\begin{itemize}

    \item GroupDRO \cite{Sagawa*2020Distributionally}  minimizes worst-case loss for distributionally robust optimization.
     \item RSC \cite{huang2020self} discards features that have higher activation to activate the remaining features appear to be applicable to out-of-domain data.
      \item MixStyle \cite{zhou2020domain} synthesizes new domains by mixing feature statistics of two features.
     \item DSU \cite{li2022uncertainty} synthesizes new domains by re-normalizing feature statistics of features with the ones drawn from a probability distribution. 
     \item LISA \cite{pmlr-v162-yao22b} selectively mixes up samples to learn an invariant predictor.
    \item Fish \cite{shi2022gradient} aligns gradients across domains by maximizing the gradient inner product.

\end{itemize}

Classification accuracy, which is the ratio of the number of correct predictions to the total number of samples is reported. 
Following the instruction of datasets, average accuracy on the test domain over 10 runs is reported for the Camelyon17 dataset; worst-group accuracy on the test domain over 3 runs is reported for the FMoW dataset. For PACS, TerraIncognita and DomainNet datasets, the averaged accuracy of the test domains over 3 runs is reported. 

Table \ref{tb:resultA} shows the results. 
Our method consistently achieves the highest average accuracy across all the datasets, outperforming SOTA  methods.
This result suggests that XDomainMix is able to train models with good domain generalization ability. 

\begin{table*}[t!]
    \centering
    \begin{subtable}{\linewidth}
        \centering
        \begin{tabular}{l|ccccc}
            \hline
            
            Method      & Camelyon17    & FMoW          & PACS            & TerraIncognita     & DomainNet      \\ \hline
            ERM         & 0.47$\pm$0.18 & 0.35$\pm$0.09 & 0.69$\pm$0.12   & 0.50$\pm$0.09   & 3.60$\pm$0.38   \\
            GroupDRO    & 0.21$\pm$0.02 & 0.40$\pm$0.04 & 0.77$\pm$0.16   & 0.43$\pm$0.05   & 2.13$\pm$0.09 \\ 
            RSC         & 0.32$\pm$0.17 & 0.55$\pm$0.14 & 42.3$\pm$12.8   & 29.9$\pm$3.02   & 17.3$\pm$1.43 \\
            MixStyle    & 0.28$\pm$0.26 & 0.36$\pm$0.05 & 0.69$\pm$0.03   & 0.38$\pm$0.09   & 3.39$\pm$0.17 \\
            DSU         & \textbf{0.07$\pm$0.02} & 0.32$\pm$0.02 & 0.33$\pm$0.04   & 9.18$\pm$1.84  & 4.61$\pm$0.24   \\
            LISA        & 0.19$\pm$0.05 & 0.29$\pm$0.05 & 0.04$\pm$0.00   & 0.13$\pm$0.01   & 0.72$\pm$ 0.06 \\
            Fish        & 3.95$\pm$3.28 & 0.47$\pm$0.02 & 0.64$\pm$0.34   & 0.34$\pm$0.04   & 2.60$\pm$0.26 \\
            XDomainMix  & 0.19$\pm$0.07 & \textbf{0.28$\pm$0.01} & \textbf{0.02$\pm$0.00}  & \textbf{0.04$\pm$0.00} & \textbf{0.11$\pm$0.02} \\ \hline
        \end{tabular}
        \caption{Representation invariance measured by distance of covariance matrix of same class representations across domains.}
    \end{subtable}
    
    \vspace{0.15cm}
    
    \begin{subtable}{\linewidth}
        \centering
        \begin{tabular}{l|ccccc}
            \hline
            Method      & Camelyon17    & FMoW          & PACS           & TerraIncognita        & DomainNet     \\ \hline
            ERM         & 1.55$\pm$0.27 & 139$\pm$50.0  & 9.89$\pm$2.09  & 12.7$\pm$2.6          & 553$\pm$26.5  \\
            GroupDRO    & 0.93$\pm$0.11 & 161$\pm$171   & 398$\pm$65.1   & 603$\pm$22.3          & 668$\pm$17.9 \\ 
            RSC         & 1.90$\pm$0.50 & 181$\pm$70.8  & 6.34$\pm$0.91  & 10.3$\pm$3.6          & 631$\pm$7.06\\
            MixStyle    & 1.67$\pm$0.91 & 110$\pm$88.9  & 6.51$\pm$1.20  & 9.10$\pm$0.56         & 563$\pm$4.68 \\
            DSU         & 3.85$\pm$1.30 & 237$\pm$136   & 16.2$\pm$5.24  & 10.6$\pm$1.5          & 567$\pm$21.1 \\
            LISA        & 1.81$\pm$1.14 & 111$\pm$15.2  & 3.02$\pm$0.47  & 9.39$\pm$0.51         & 520$\pm$11.9\\
            Fish        & 5.68$\pm$1.81 & 251$\pm$45.3  & 9.08$\pm$5.38  & 9.37$\pm$1.59         & 567$\pm$13.2 \\
            XDomainMix  & \textbf{0.90$\pm$0.28} & \textbf{109$\pm$11.7}  & \textbf{2.10$\pm$0.21}   & \textbf{8.22$\pm$1.05} & \textbf{504$\pm$15.8}\\ \hline
        \end{tabular}
        \caption{Prediction invariance measured by variance of risk across domains. The results are reported in the unit of 1e-3.}
    \end{subtable}
\caption{Results of model invariance.}
\label{tb:invariance_result}
\end{table*}

\begin{table*}[t!]
    \centering
    \begin{tabular}{l|ccccc}
    \hline
    
         Method     & Camelyon17       & FMoW           & PACS          & TerraIncognita & DomainNet  \\\hline
         MixStyle   & 6.65$\pm$0.18    & 8.58$\pm$0.50    & 3.11$\pm$0.34            & 3.20$\pm$0.26  & 2.81$\pm$0.11\\
         DSU        & 26.77$\pm$2.53   & 20.93$\pm$1.87 &  6.97$\pm$0.03           & 10.85$\pm$0.32 & 5.75$\pm$0.01\\
         XDomainMix & \textbf{38.82$\pm$0.28} & \textbf{34.06$\pm$0.11} & \textbf{14.82$\pm$0.11} & \textbf{14.36$\pm$0.12} & \textbf{10.27$\pm$0.02} \\\hline
    \end{tabular}
    \caption{Divergence of the augmented feature and original feature measured by MMD in the unit of 1e-2.} 
    \label{tab:divergence}
\end{table*}

\subsection{Model Invariance}
One advantage of XDomainMix is that is able to learn invariance across training domains.
We quantify the invariance in terms of representation invariance and predictions invariance.
Representation invariance refers to the disparity between representations of the same class across different domains. The distance between second-order statistics (covariances) \cite{sun2016deep} can be used to measure representation invariance. Prediction invariance considers the variation in predictions across different domains.
We employ risk variance \cite{pmlr-v162-yao22b} which measures how similar the model performs across domains.

Let $\{Z^{(i)}_j|y^{(i)}_j=y_c\}$ be the set of representations of class label $y_c$ from domain $i$. We use $C^{(i)}_{y_c}$ to denote the covariance matrix of the representations.
Given the set of class labels $\mathcal{Y}$ and the set of training domains $\mathcal{D}_S$, the measurement result is given by $
    \frac{1}{|\mathcal{Y}||\mathcal{D}_S|}
\sum_{y_c\in\mathcal{Y}}\sum_{i,i' \in \mathcal{D}_S} ||C^{(i)}_{y_c}- C^{(i')}_{y_c}||^2_F
$. $||\cdot||^2_F$ denotes the squared matrix Frobenius norm.
A smaller distance suggests that same-class representations across domains are more similar.

Let $R^i $ be the loss in predicting the class labels of inputs from domain $i$.
The risk variance is given by the variance among training domains, $\text{Var}\{R^1, R^2, ..., R^{|\mathcal{D}_S|}\}$. 
Lower risk variance suggests a more consistent model performance across domains.

 Table \ref{tb:invariance_result} shows the results.  We see that our method has the smallest covariance distance in the FMoW, PACS, TerraIncognita, and DomainNet dataset, and the second-smallest in the Camelyon17 dataset. The results indicate that the representations of the same class learned by our method have the least divergence across domains.
 Additionally, XDomainMix has the lowest risk variance, suggesting that it is able to maintain consistent performance in predictions across domains. 
Overall, the results demonstrate that our approach is able to learn invariance at both the representation level and the prediction level.

\subsection{Diversity of Augmented Features}
To show that XDomainMix can generate more diverse features, we measure the distance between the original and augmented features using maximum mean discrepancy (MMD). A higher MMD suggests that the distance between the original and augmented features is further. 
The same set of original features is used to ensure fairness and comparability of the measurement result.  Average and standard deviation over three runs are reported. 
Table \ref{tab:divergence} shows the results. Features augmented by XDomainMix consistently have the highest MMD compared to MixStyle and DSU which are two SOTA feature augmentation methods. 
This suggests that the features augmented by XDomainMix exhibit the most  deviation from the original features, leading to a more varied augmentation. 
Visualization of  sample images reconstructed from augmented features  are given in Supplementary.

\subsection{Experiments on the Identified Features}
\label{sc:feature_decomposition_ana}
In this set of experiments, we demonstrate that XDomainMix is able to identify features that are important for class and domain prediction.
We evaluate the model performance for class or domain prediction after eliminating those features with the highest importance score computed in Equations \ref{eq:class_im} and \ref{eq:domain_im}. A decrease in accuracy suggests that the features that have been removed are important for the predictions. 

For comparison, we implement two alternative selection strategies:  a random method that arbitrarily selects features to remove, and a gradient norm approach, where features are chosen for removal based on the magnitude of the gradient in the importance score computation.
Samples in the validation set of PACS dataset are used in this experiment.

\begin{figure}[t!]
     \centering
     \begin{subfigure}{0.495\linewidth}
     \centering
     \includegraphics[width=\linewidth]{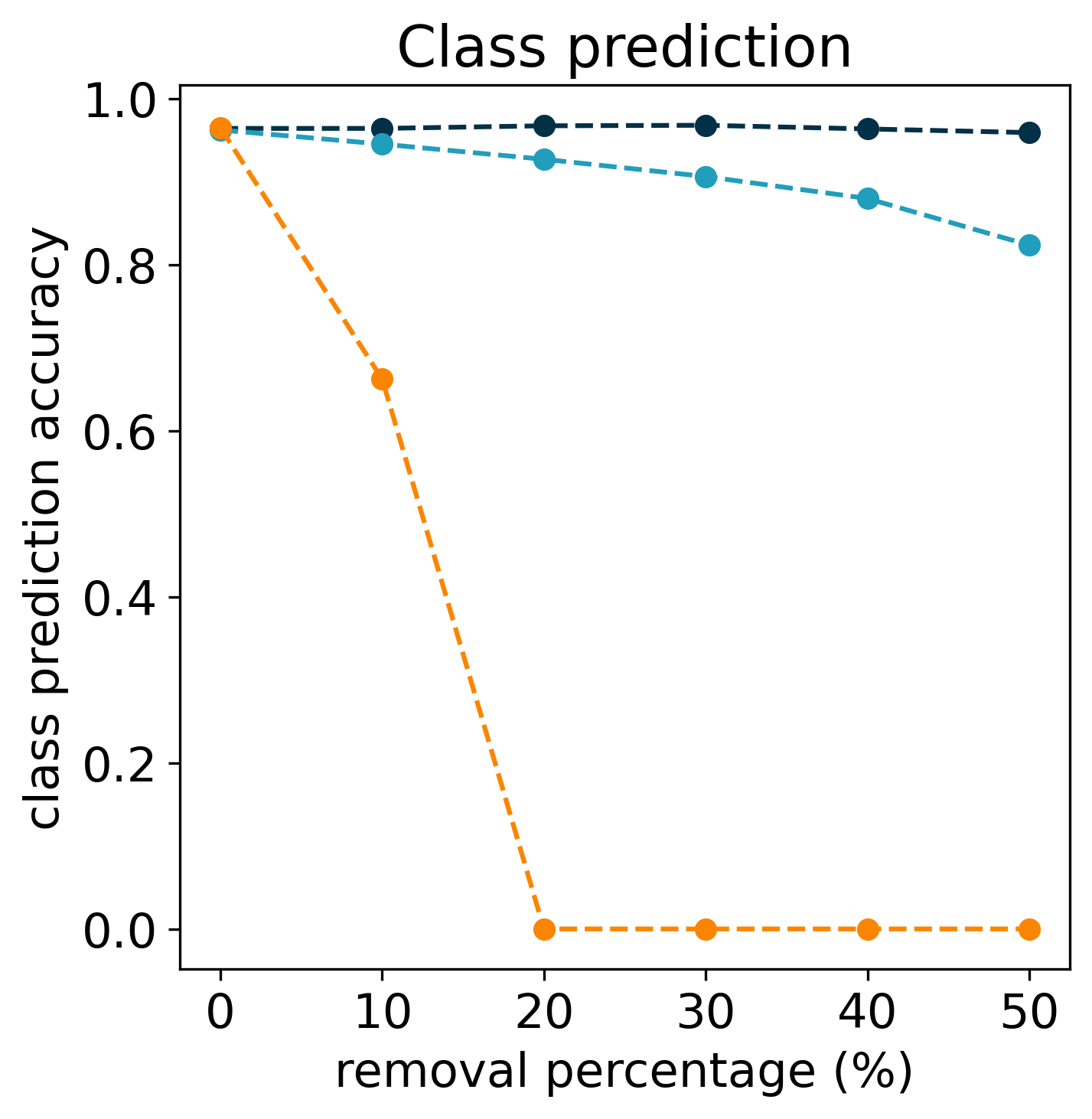}\\
     \end{subfigure}
     \begin{subfigure}{0.495\linewidth}
     \centering
        \includegraphics[width=\linewidth]{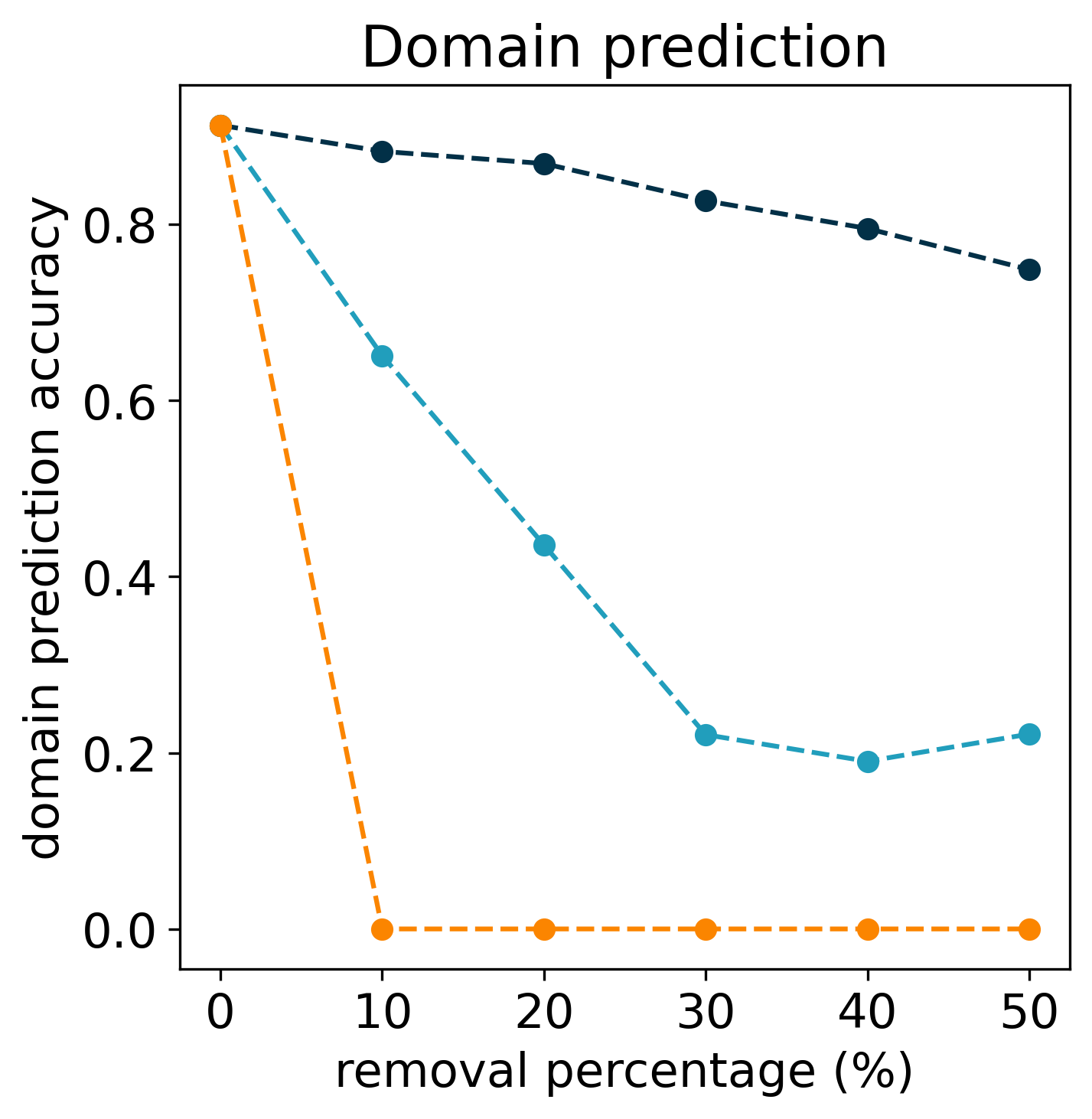}\\
     \end{subfigure}
     \includegraphics[width=0.55\linewidth]{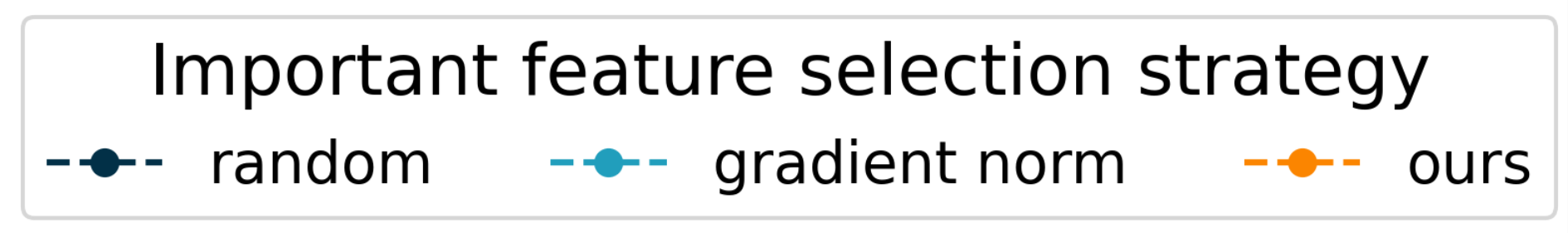}
     \caption{Prediction accuracy after removing x\% of features with the highest importance scores.}
     \label{fig:feature_im}
\end{figure}

Figure \ref{fig:feature_im} shows the results.
Our method shows the largest drop in both class prediction and domain prediction accuracies compared to random removal and gradient norm methods. This indicates that XDomainMix is able to identify features that are specific to the domain and class effectively.

\begin{figure}[t!]
    \centering
    \begin{subfigure}{\linewidth}
    \centering
    \includegraphics[width=\linewidth]{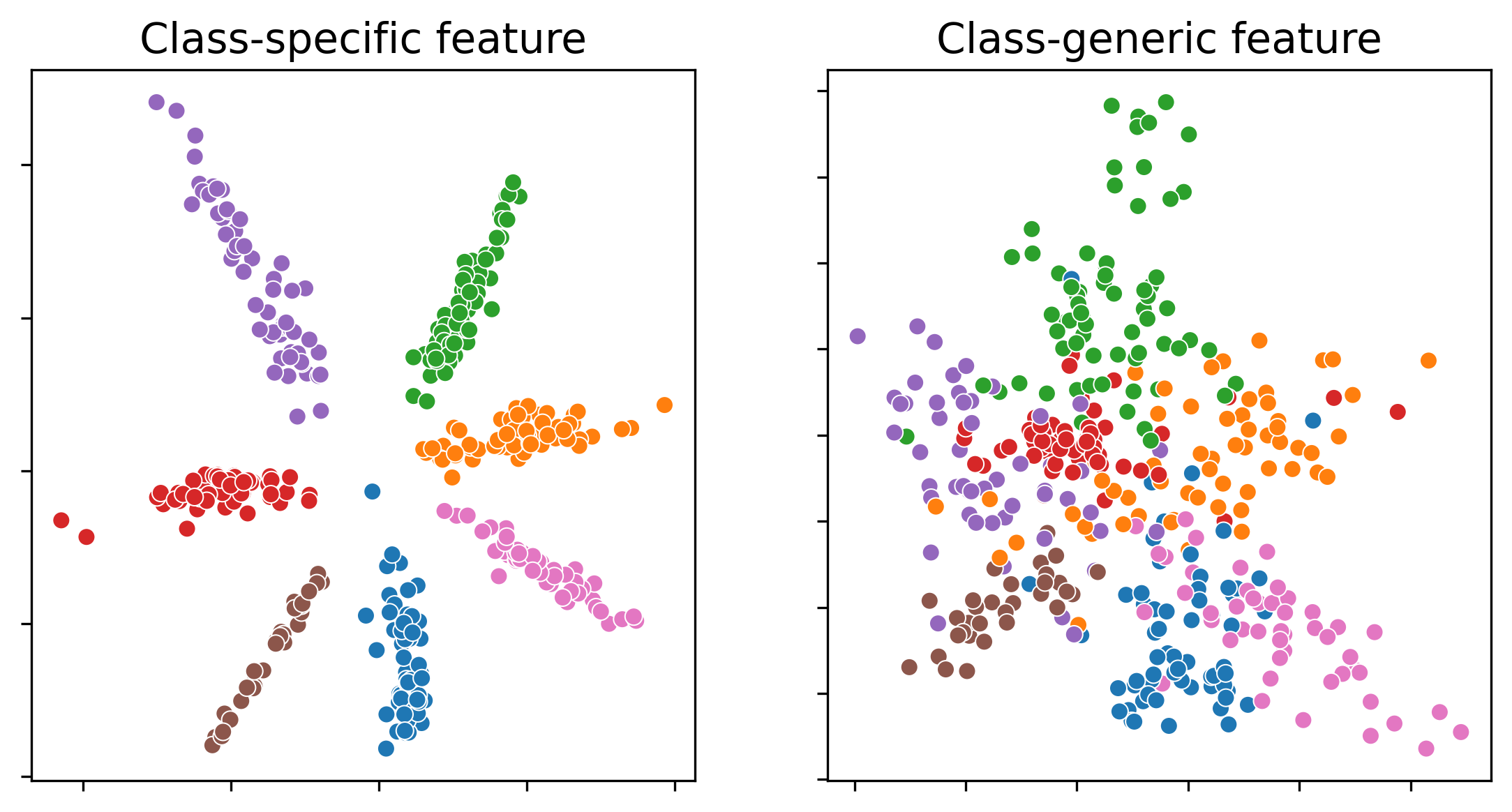}
    \caption{Class-related features  }
   
    \end{subfigure}
    \begin{subfigure}{\linewidth}
    \centering
    \includegraphics[width=\linewidth]{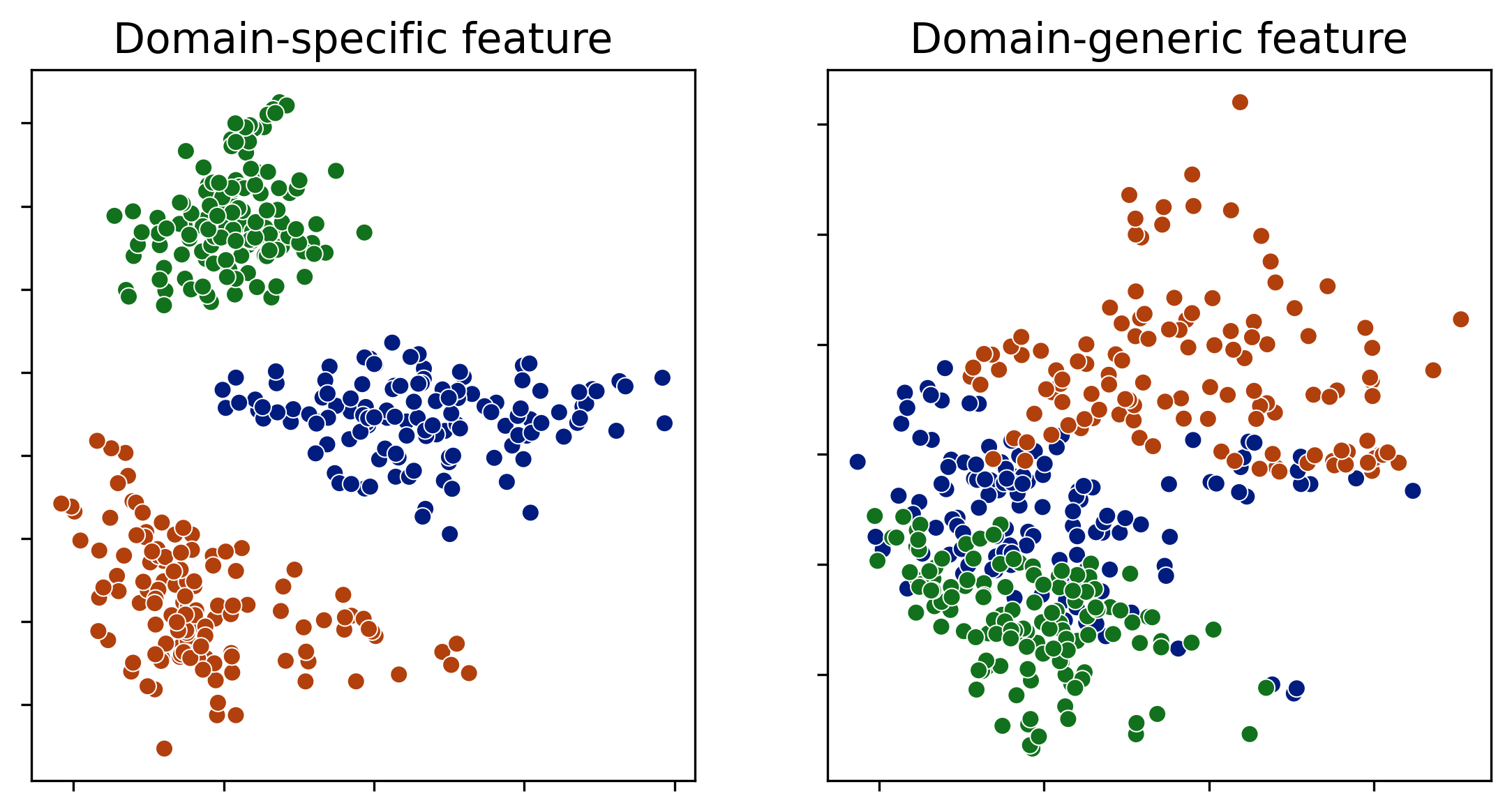}
    \caption{Domain-related features }

    \end{subfigure}
    \caption{Visualization of features from different classes/domains, indicated by the different colors.}
  \label{fig:classdomain_feature} 
\end{figure}

To visualize the extracted features, we 
 map the high dimensional feature vectors obtained by $f$ to a lower dimensional space,
This transformation is carried out using two linear layers, as described in \cite{zhang2021unleashing}.
Figure \ref{fig:classdomain_feature}(a) provides a  visualization for the model that is trained on the PACS dataset, with Art as the unseen domain.  
We see that the features identified as class-specific are well separated by class. This is in contrast to the features that are generic across classes, which are not as clearly delineated.

Similarly, we also visualize the extracted domain-specific and domain-invariant features. As shown in Figure \ref{fig:classdomain_feature}(b),  the domain-specific features are noticeably better separated compared to the domain-invariant features.

\subsection{Ablation Study}
To understand the contribution of each component in the augmentation, we perform ablation studies on Camelyon17 and FMoW datasets. Table \ref{tab:ablation} shows the result. 
Compared to the baseline, mixing class-specific domain-specific feature components ($Z_{c,d}$) only, or mixing class-generic domain-specific feature components ($Z_{\neg c,d}$) only in the augmentation can improve the performance.
This suggests that by manipulating domain-specific feature components, models that are better at domain generalization can be learned. 
Mixing $Z_{\neg c,d}$ leads to greater improvement, indicating that enriching diversity by content from other classes is more helpful than simply intra-class augmentation.

Augmenting both $Z_{c,d}$ and $Z_{\neg c,d}$ does not consistently lead to performance improvement, possibly due to dataset-specific characteristics. 
Probabilistically discarding $Z_{c,d}$ seems to encourage the model to use less domain-specific information and exploit less activated features in prediction, which improves the domain generalization performance.

\begin{table}[ht!]
 
    \centering
    \begin{tabular}{ccc|cc}\hline
         mix & mix & discard & Camelyon17    & FMoW \\
         $Z_{c,d}$ &  $Z_{\neg c,d}$  &  $Z_{c,d}$ \\ \hline
         \multicolumn{3}{l|}{}                                   & 70.3$\pm$6.4 & 32.3$\pm$1.3 \\
         $\checkmark$  &                     &                   & 78.3$\pm$5.5  & 32.9$\pm$2.2\\
                       & $\checkmark$        &                   & 79.1$\pm$6.0  & 33.6$\pm$1.1\\
         $\checkmark$  & $\checkmark$        &                   & 79.6$\pm$7.0  & 31.9$\pm$0.4\\
         $\checkmark$  & $\checkmark$        & $\checkmark$      & \textbf{80.9$\pm$3.2} & \textbf{35.9$\pm$0.8}\\ \hline
    \end{tabular}
    \caption{Ablation study.}
    \label{tab:ablation}
\end{table}

\section{Conclusion and Future Work}
In this work, we have developed a feature augmentation method to address the domain generalization problem. Our approach aims to enhance data diversity within the feature space for learning models that are invariant across domains by mixing domain-specific components of features from different domains while retaining class-related information. We have also probabilistically discarded domain-specific features to discourage the model from using such features for their predictions, thereby achieving good domain generalization performance. Our experiments on multiple datasets demonstrate the effectiveness of the proposed method.

While our method presents a promising approach to solving the domain generalization problem, there are several limitations. 
Our method needs more than one training domain to perform cross-domain feature augmentation.
Our method assumes that the datasets across different domains share the same label space and similar class distributions, and its performance may be affected if this is not the case.
Our method is mainly empirically validated, and a theoretical analysis or guarantee of its performance is still lacking. Further research is needed to provide a deeper theoretical understanding of the proposed method and its performance bounds.

\section*{Acknowledgements}
This research/project is supported by the National Research Foundation, Singapore under its AI Singapore Programme (AISG Award No: AISG-GC-2019-001-2B). We thank Dr Wenjie Feng for the helpful discussions.

\bibliographystyle{named}
\bibliography{ref}

\clearpage
 
\appendix

\section{Comparison with Sharpness-aware Methods}
Apart from learning invariance across domains, learning a flat minimum is another approach to improve domain generalization, as recent work suggests that flat minima bring better generalization than sharp minima. As a result, several domain generalization works seek flat minima by optimization that leads to flatter loss landscapes. 

Here we compare the performance of XDomainMix with two sharpness-aware methods: 
\begin{itemize}
    \item SAM \cite{foret2021sharpnessaware} seeks parameters that lie in neighborhoods that have uniformly low loss.
    \item SAGM \cite{wang2023sharpness} aligns the gradient direction between the SAM loss and the empirical risk. 
\end{itemize}

The results are shown in Table \ref{tab:dg_sharpness}. We follow the same experiment and reporting protocol as that in Table 1. 
XDomainMix outperforms SAM and SAGM on Camelyon17 and FMoW datasets. On the PACS and TerraIncognita datasets, XDomainMix comes as a close second to SAGM.
The result is expected as the superiority of considering sharpness in domain generalization has been empirically demonstrated. 

It is worth mentioning that sharpness-aware methods can be easily incorporated into XDomainMix to learn a flatter minimum. We see that XDomainMix+SAM achieves better performance on Camelyon17, FMoW, PACS, and DomainNet datasets, indicating that incorporating sharpness-related strategies can further boost the performance of XDomainMix.

\begin{table*}[!ht]
    \centering
    \begin{tabular}{l|lllll}\hline
    Method          &  Camelyon17                & FMoW                & PACS           & TerraIncognita    & DomainNet\\ \hline
    SAM             &  75.8$\pm$5.9              & 35.4$\pm$1.4        & 85.8$\pm$0.2   & 43.3$\pm$0.7      & 44.3$\pm$0.0\\
    SAGM            &  80.0$\pm$2.9              & 31.0$\pm$1.6        & \textbf{86.6$\pm$0.2} & \textbf{48.8$\pm$0.9}  &\textbf{45.0$\pm$0.2} \\
    XDomainMix      & \textbf{80.9$\pm$3.2}      & \textbf{35.9$\pm$0.8} & 86.4$\pm$0.4   & 48.2$\pm$1.3      & 44.0$\pm$0.2 \\ \hline
    XDomainMix+SAM  & \textbf{81.4$\pm$3.1}      & \textbf{36.1$\pm$1.3} & \textbf{86.7$\pm$0.4}   & 44.4$\pm$0.4      & \textbf{45.1$\pm$0.1}  \\ \hline
    \end{tabular}
    \caption{Domain generalization performance of XDomainMix compared with sharpness-aware methods.}
    \label{tab:dg_sharpness}
\end{table*}

\section{Additional Results on Experiments of the Identified Features}
\begin{figure}[!ht]
     \centering
     \begin{subfigure}{\linewidth}
     \centering
     \includegraphics[width=0.495\linewidth]{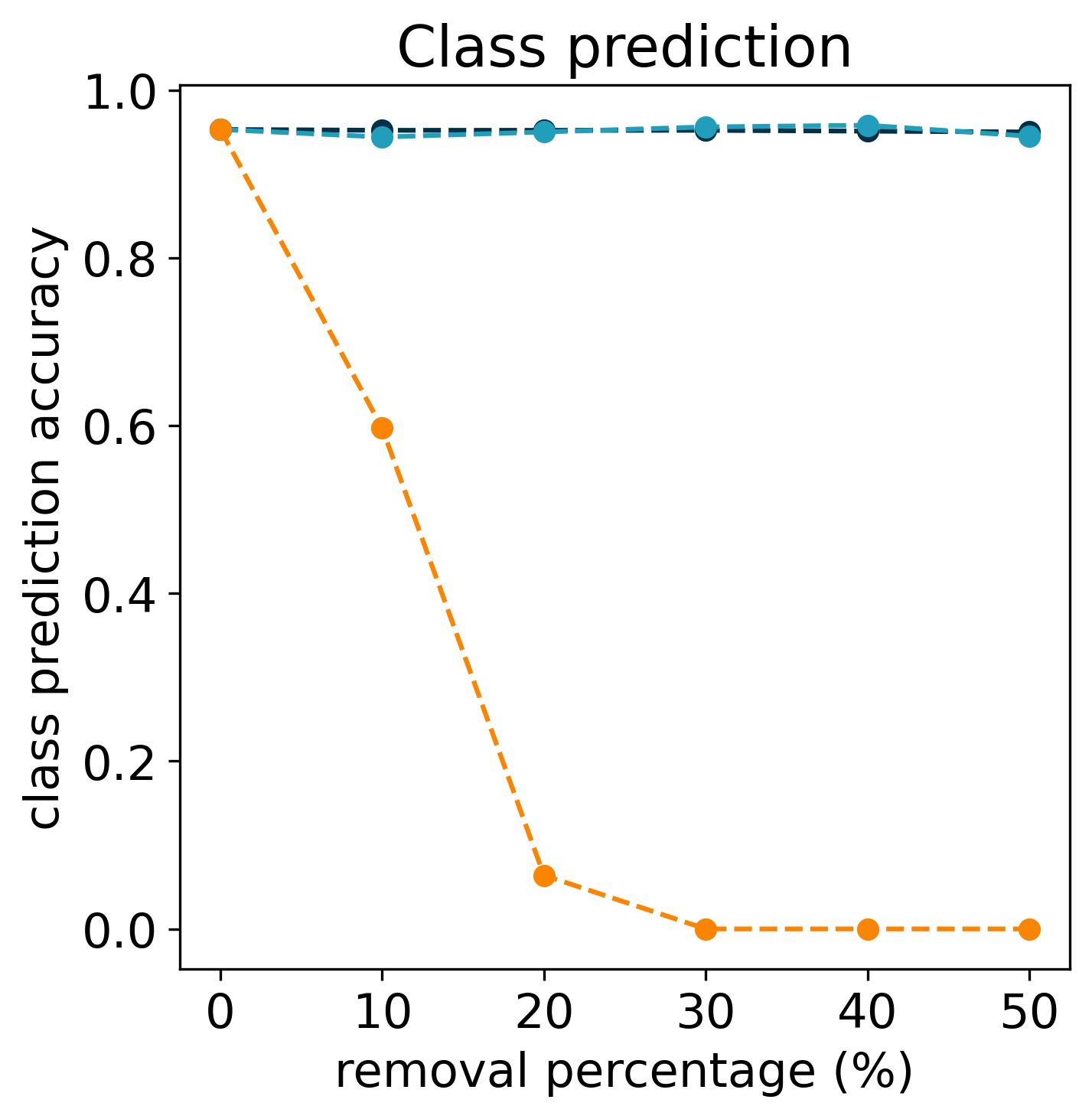}
     \includegraphics[width=0.495\linewidth]{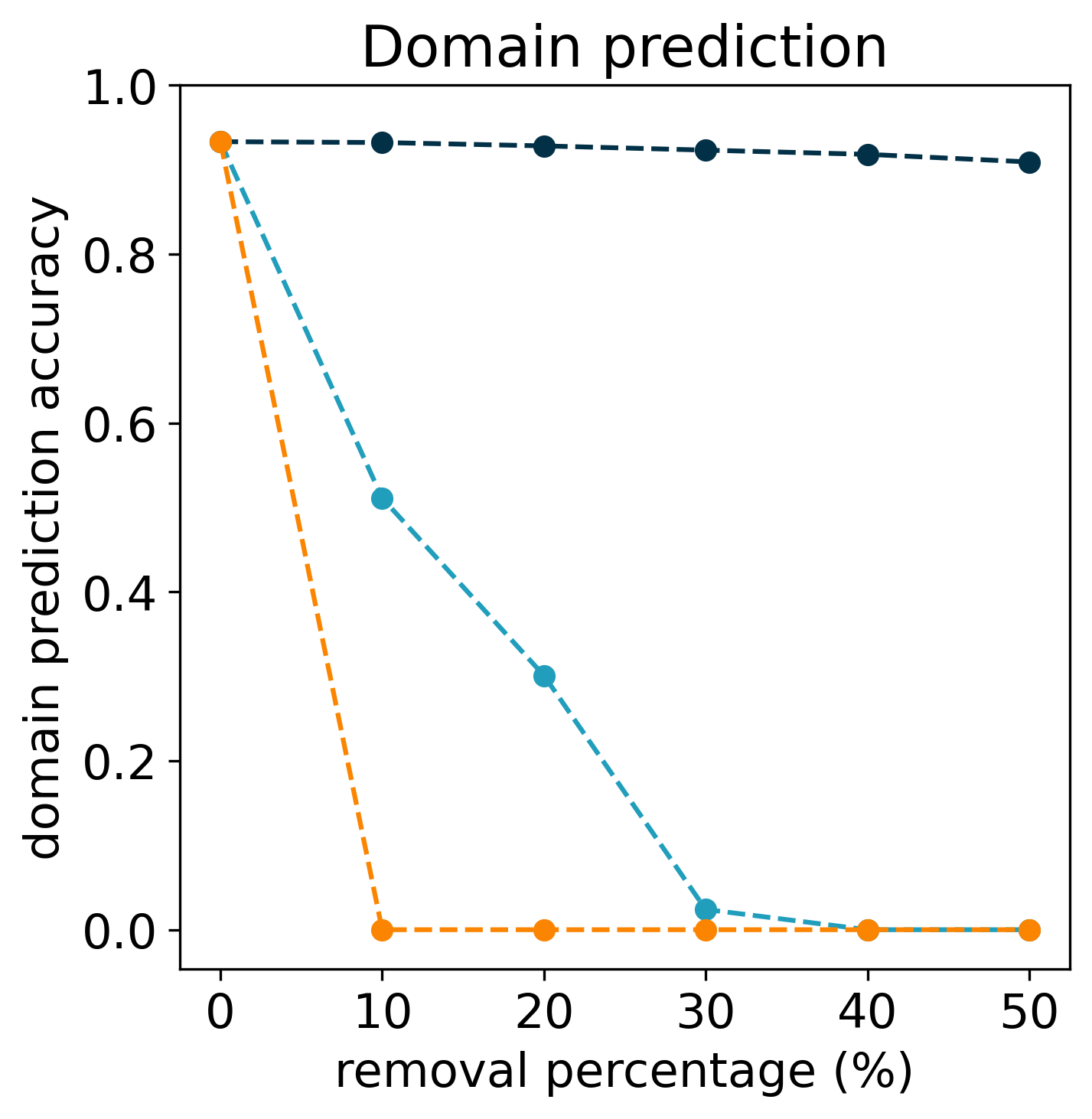}
     \subcaption{Camelyon17}
     \end{subfigure}

     \begin{subfigure}{\linewidth}
     \centering
     \includegraphics[width=0.495\linewidth]{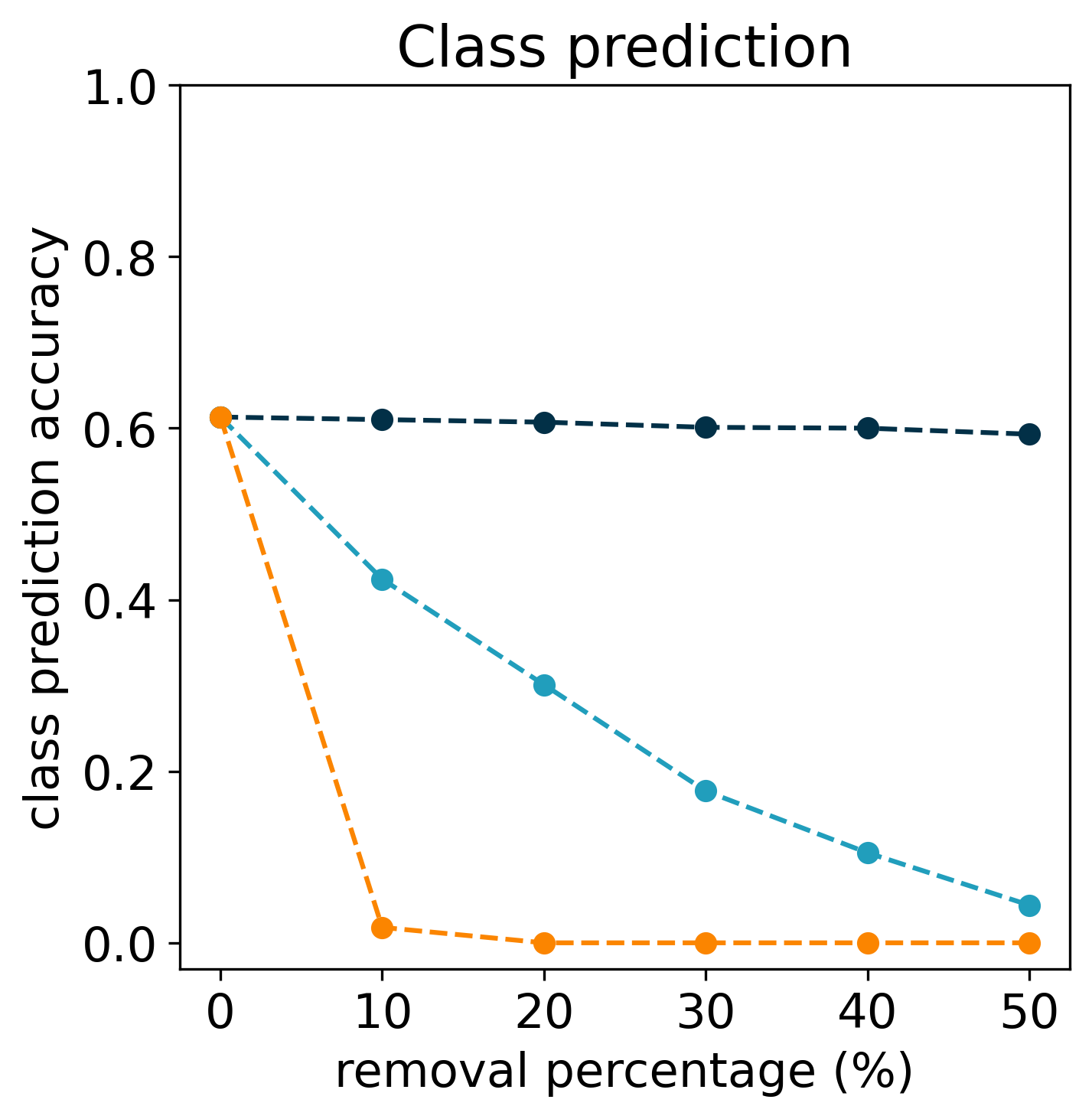}
     \includegraphics[width=0.495\linewidth]{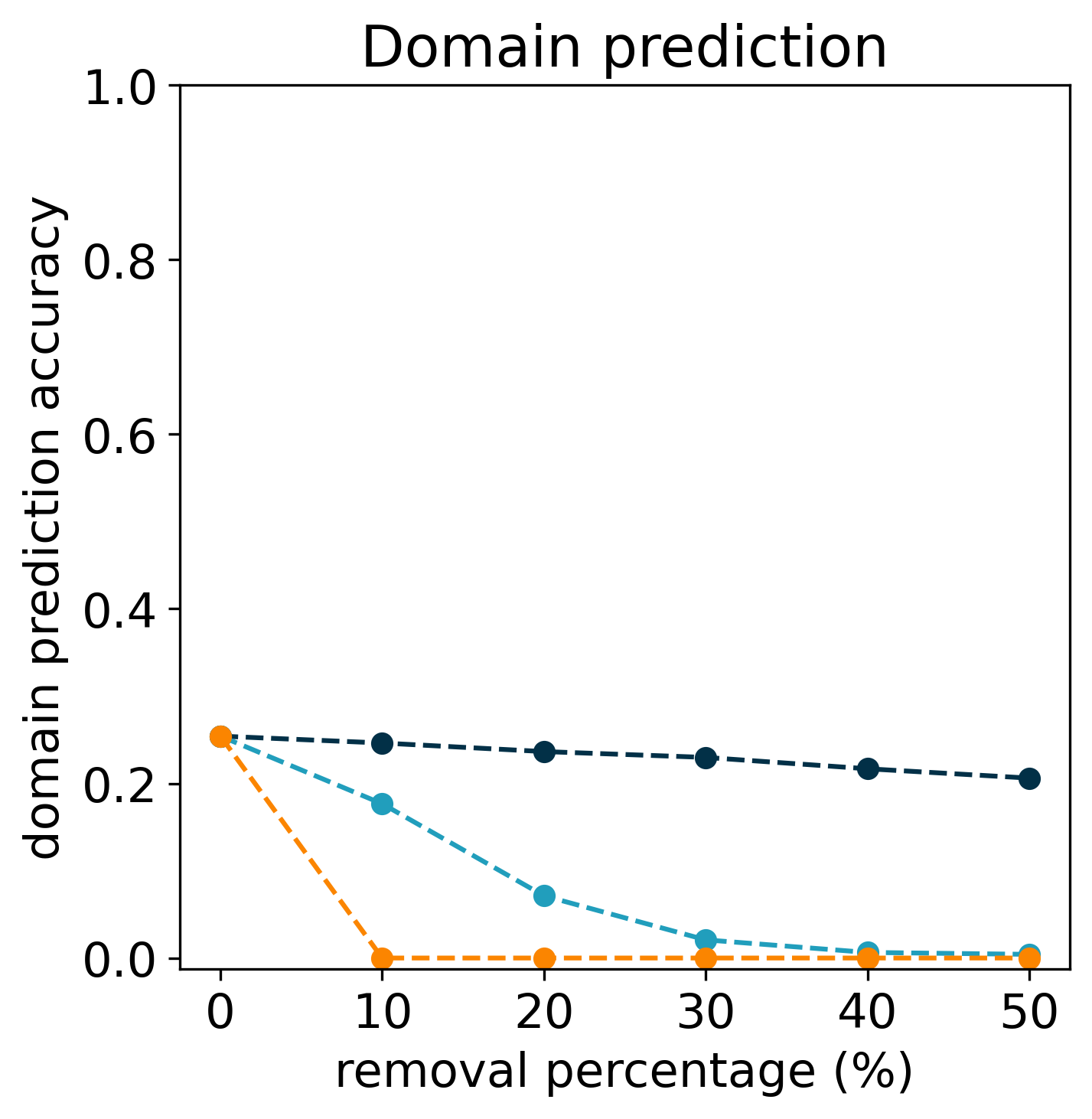}
     \subcaption{FMoW}
     \end{subfigure}

     \begin{subfigure}{\linewidth}
     \centering
     \includegraphics[width=0.495\linewidth]{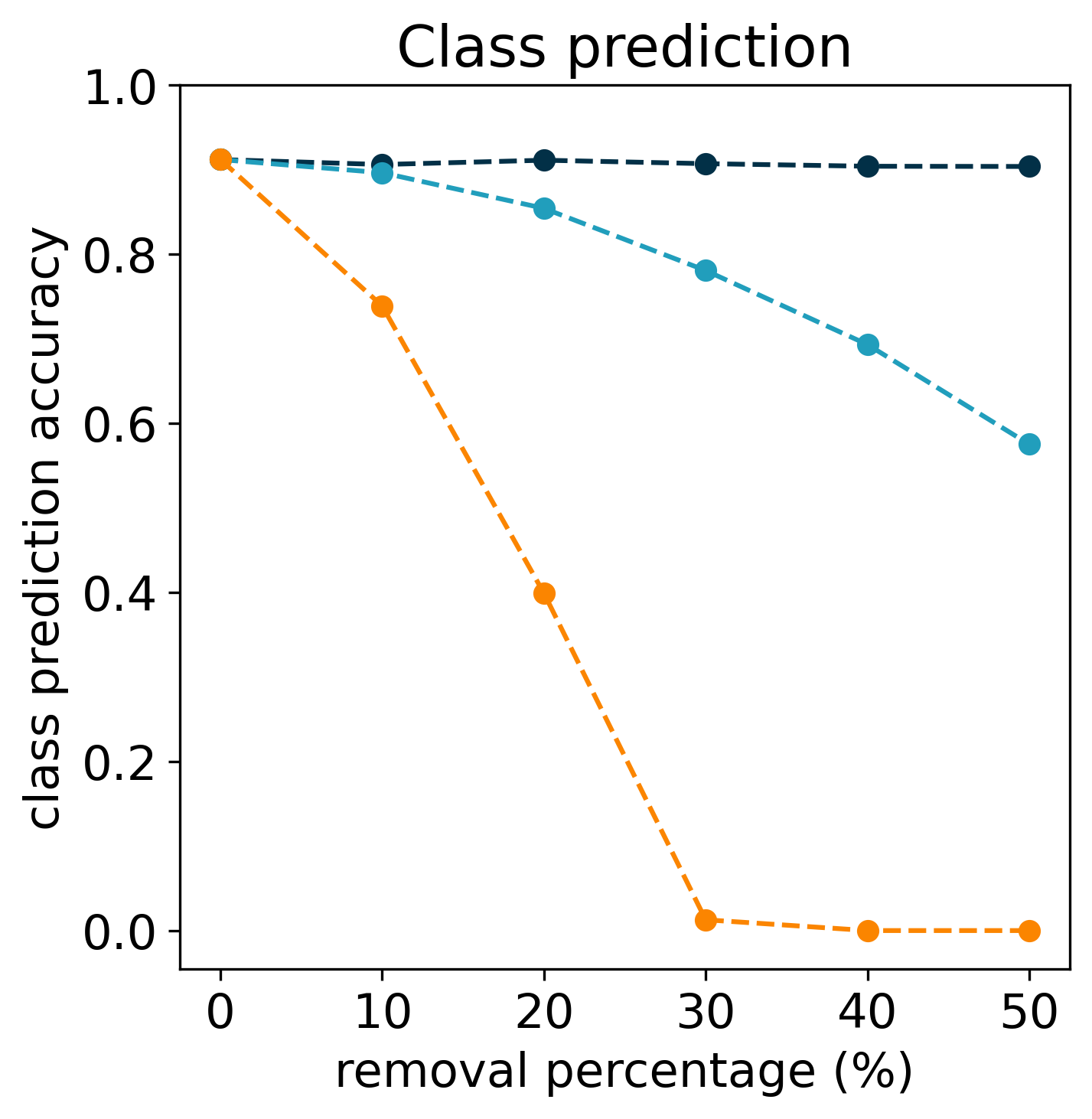}
     \includegraphics[width=0.495\linewidth]{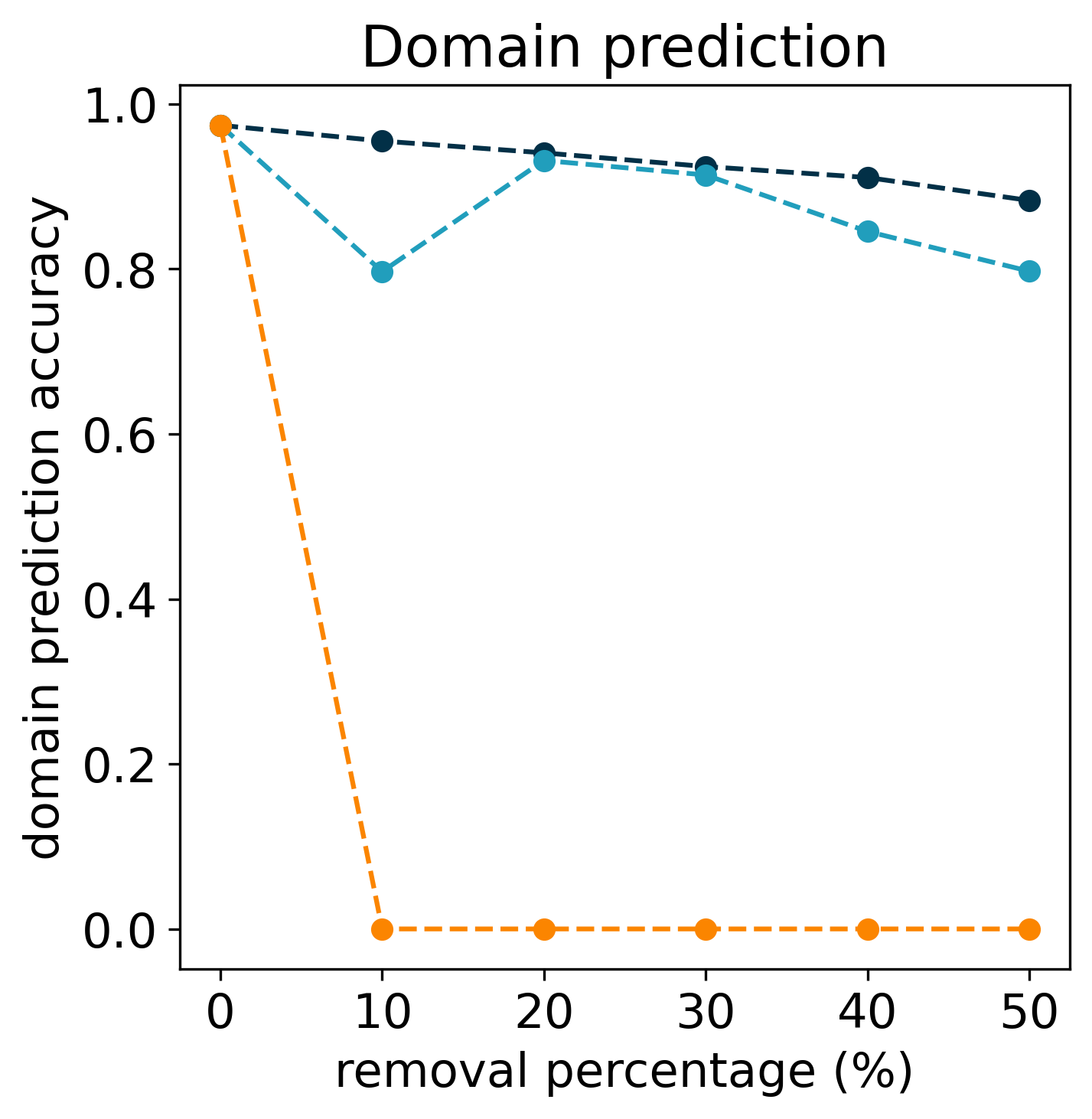}
     \subcaption{TerraIncognita}
     \end{subfigure}

     \begin{subfigure}{\linewidth}
     \centering
     \includegraphics[width=0.495\linewidth]{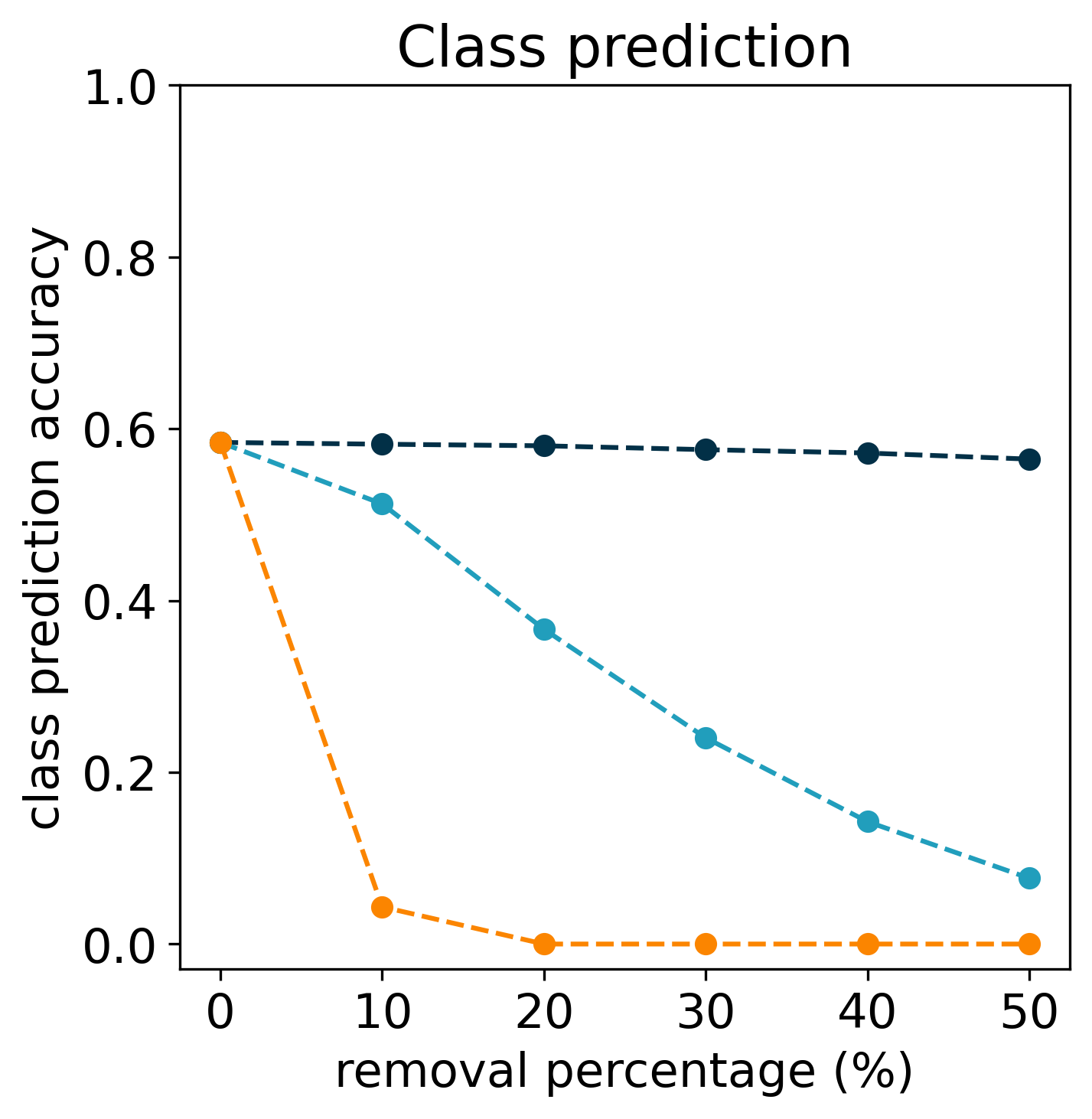}
     \includegraphics[width=0.495\linewidth]{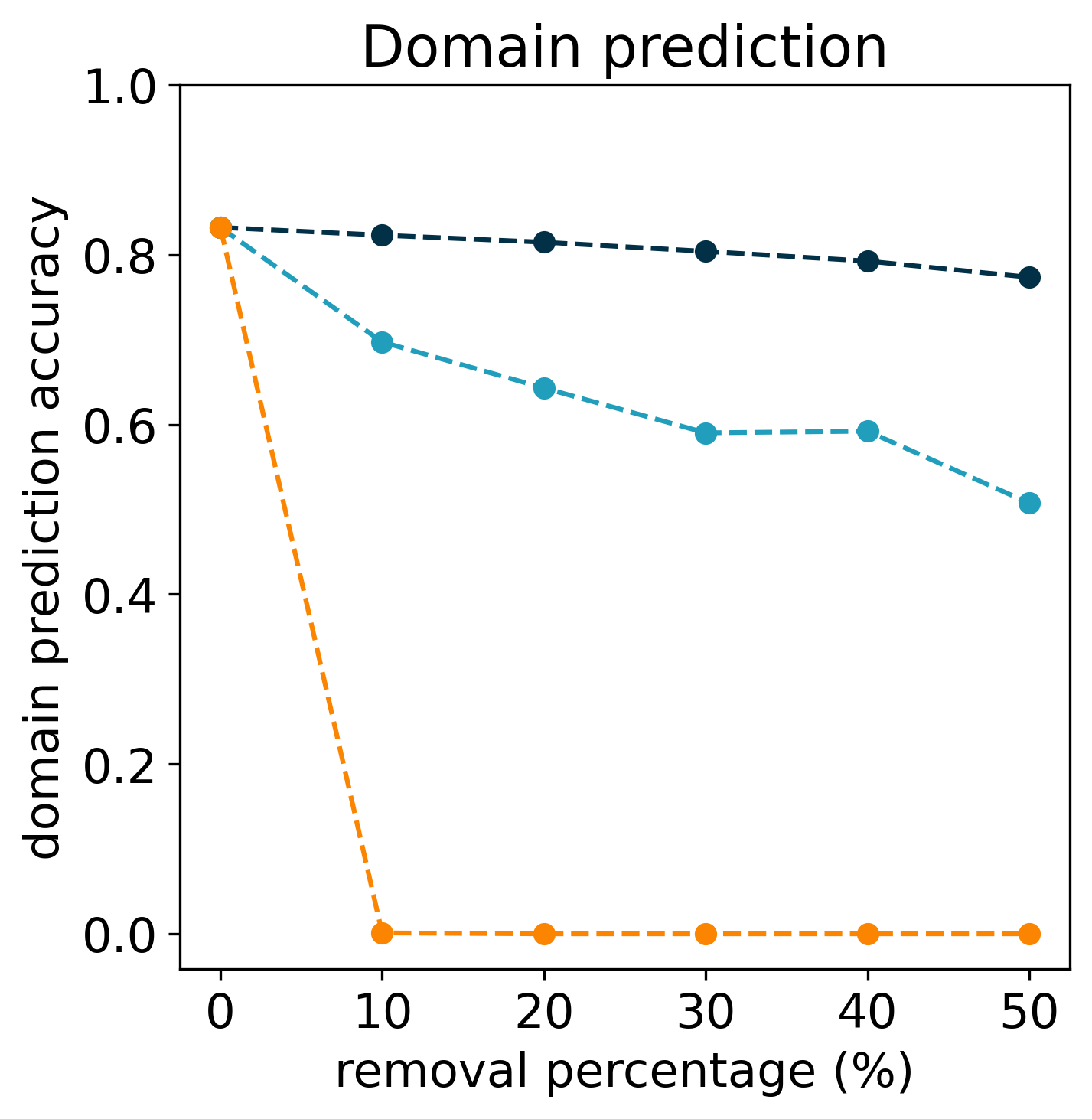}
     \subcaption{DomainNet}
     \end{subfigure}
     
     \includegraphics[width=0.55\linewidth]{figures/legend_remove.png}
     \caption{Prediction accuracy after removing x\% of features with the highest importance scores on additional datasets.}
     \label{fig:feature_im_add}
     
\end{figure}

We present the results of eliminating features with the highest importance score computed in Equations 1 and 2 on other datasets in Figure \ref{fig:feature_im_add}. As that we see on the PACS dataset, XDomainMix outperforms random selection and gradient norm methods by exhibiting the most significant decline in both class and domain prediction accuracies on other datasets. This highlights its effective identification of class and domain-specific features.

\section{Scalability}
Large-scale foundation models have emerged as a prominent trend. 
XDomainMix can be applied to any model architecture that allows for the decomposition of its features, including larger and more complex models like Vision Transformer \cite{dosovitskiy2020image}. 
We applied the ViT-B/16 CLIP model \cite{radford2021learning} to XDomainMix by using its image encoder to extract features and fine-tuning the classifier. Feature augmentation is performed on the features extracted by the image encoder. 

We compared with CLIP's zero-shot prediction and ERM fine-tuning of the classifier. We use the prompt template ``a photo of a \{class name\}'' for zero-shot prediction. For ERM and XDomainMix fine-tuning, the image encoder is frozen, and only the classifier is updated.
We follow the same experiment and reporting protocol as that in Table 1 for finetuning.
Table \ref{tab:dg_clip} shows the result. 
On the Camelyon17 dataset, CLIP's zero-shot prediction achieves fair performance. Finetuning the classifier further improves the performance, and XDomainMix gives better result than ERM.
On the FMoW dataset, XDomainMix finetuning gives the best performance. CLIP's subpar zero-shot prediction suggests that the image encoder may not be optimal for FMoW. While finetuning enhances performance, it falls short of achieving the levels seen in Table 1.

\begin{table}[!ht]
    \centering
    \begin{tabular}{l|cc}\hline
    Method          &  Camelyon17     & FMoW            \\ \hline
    zero-shot       & 68.2            & 12.9            \\
    ERM             & 86.4$\pm$0.3    & 26.6$\pm$0.4    \\
    XDomainMix      & \textbf{86.6$\pm$0.3} & \textbf{26.9$\pm$0.2}   \\\hline
    \end{tabular}
    \caption{Domain generalization performance with ViTB/16 CLIP.}
    \label{tab:dg_clip}
\end{table}

\section{Visualization of Augmented Features} 
\label{ap:visualization}
\begin{figure}[!ht]
        \centering
        \begin{subfigure}{\linewidth}
            \begin{minipage}{0.3\linewidth}
            \centering
            Original  \\
            \includegraphics[width=\linewidth]{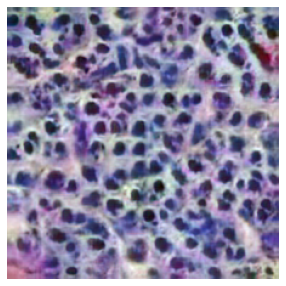}\\
            \includegraphics[width=0.95\linewidth]{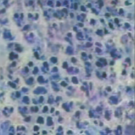}\\
           
        \end{minipage}\hfill
        \begin{minipage}{0.3\linewidth}
            \centering
             DSU \\
            \includegraphics[width=\linewidth]{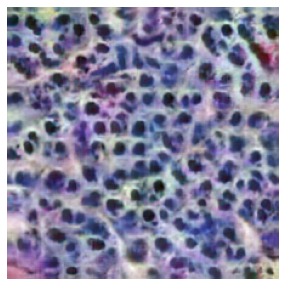}\\
            \includegraphics[width=0.95\linewidth]{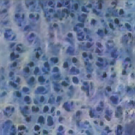}\\
           
        \end{minipage}\hfill
        \begin{minipage}{0.3\linewidth}
            \centering
            XDomainMix \\
            \includegraphics[width=\linewidth]{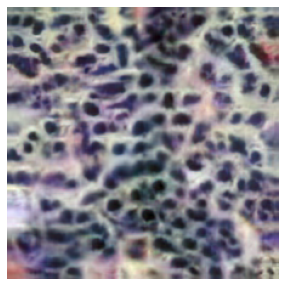}\\
            \includegraphics[width=0.95\linewidth]{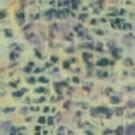}\\
        \end{minipage}
        \caption{Normal tissues}
        \end{subfigure}
        \begin{subfigure}{\linewidth}
            \begin{minipage}{0.3\linewidth}
            \centering
            Original 
            \includegraphics[width=\linewidth]{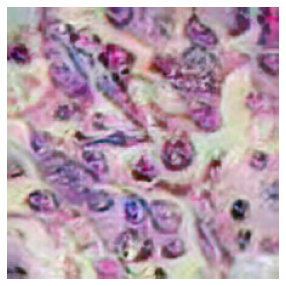}\\
            \includegraphics[width=\linewidth]{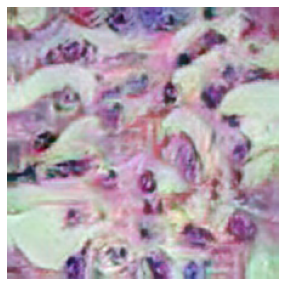}\\
            
        \end{minipage}\hfill
        \begin{minipage}{0.3\linewidth}
            \centering
            DSU 
            \includegraphics[width=\linewidth]{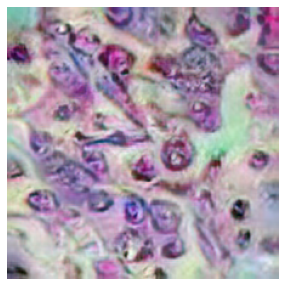}\\
            \includegraphics[width=\linewidth]{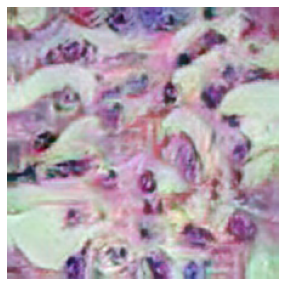}\\
            
        \end{minipage}\hfill
        \begin{minipage}{0.3\linewidth}
            \centering
             XDomainMix 
            \includegraphics[width=\linewidth]{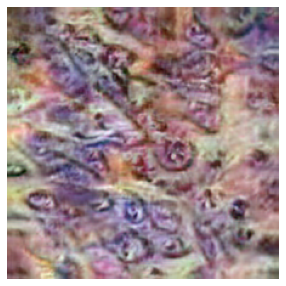}\\
            \includegraphics[width=\linewidth]{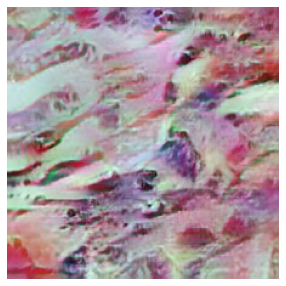}\\
           
        \end{minipage}
        \caption{Tumor tissues}
        \end{subfigure}
          
        \caption{Visualization of images reconstructed using augmented features obtained from DSU and XDomainMix. Features from XDomainMix result in samples that are more diverse than DSU method.}
        \label{fig:fig5}
    \end{figure}

We visualize the augmented features by employing a pre-trained autoencoder \cite{Huang_2017_ICCV} \footnote{Weights are from https://github.com/naoto0804/pytorch-AdaIN.} to map these features back in the input space. Figure \ref{fig:fig5} shows the reconstructed images using both the original and augmented features generated from the Camelyon17 dataset.

Cell nuclei and the general structural features of the tissue are highlighted by the stain. In general, tumor cells are larger than normal cells \cite{bandi2018detection}. For both classes, XDomainMix is able to generate augmented features with greater diversity, while DSU's augmented features show only limited differences.  In addition, XDomainMix also preserves class semantics as no large cells are included in the generated results for the normal class.
This demonstrates the effectiveness of using XDomainMix for diverse feature augmentation.

\begin{figure}[!ht]
    \centering
    \includegraphics[width=0.8\linewidth]{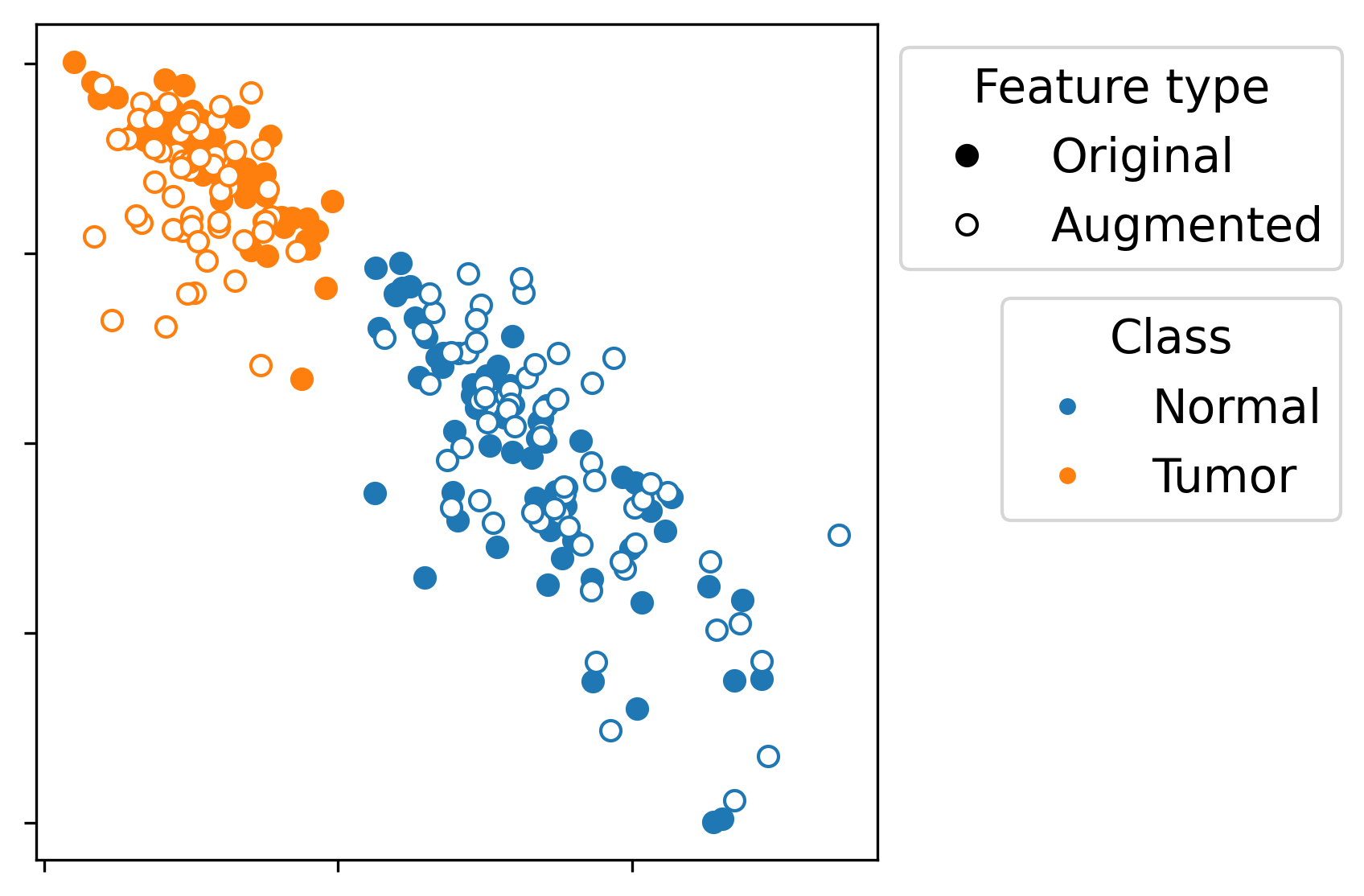}
     \caption{Visualization of original and augmented features.}
    \label{fig:pacs_aug}
\end{figure}

Additionally, we also visualize the XDomainMix augmented features in a lower dimensional space (see Figure \ref{fig:pacs_aug}). The augmented features clearly lie in the same cluster formed by the original features of their respective classes, indicating that  XDomainMix is able to preserve class-specific information.
    
\section{Alternative Design of XDomainMix}
\paragraph{Sample selection in the interpolation of class-generic domain-specific Feature Component}
To augment the feature $Z$ extracted from input $x$, we randomly sample two inputs $x_i$ and $x_j$ whose domains are different from $x$. $x_i$ has the same class label as $x$ and is used to interpolate $Z_{c,d}$. $x_j$ is from a different class and is used to interpolate $Z_{\neg c,d}$.
We select $x_j$ from a different class and a different domain so that the augmented feature has greater diversity, compared to samples from the same class or domain.

We compare the maximum mean discrepancy (MMD) between the original and augmented features on the Camelyon17 dataset when different $x_j$ selection strategies are used in the interpolation of $Z_{\neg c, d}$. A higher MMD suggests that the features are more diverged. 
Table \ref{tab:divergence_2} shows the result. 
When same-class or same-domain inputs are sampled, the augmented features always have a lower divergence. 
Selecting samples from different classes and different domains results in the highest MMD, which implies the greatest diversity.

\begin{table}[!ht]
    \centering
    \begin{tabular}{l|c}
    \hline
             Sample used  & MMD (1e-2) \\\hline
             same as $x_i$  & 35.94$\pm$0.11\\
             same class different domain (not $x_i$)    & 36.00$\pm$0.03\\
             different class same domain & 38.24$\pm$0.23 \\
               different class different domain &\textbf{38.82$\pm$0.28} \\\hline
    \end{tabular}
    \caption{Feature divergence of different sample selection in the augmentation of $Z_{\neg c, d}$ on Camelyon17 dataset}
    \label{tab:divergence_2}
\end{table}

\paragraph{Training domain classifier with augmented features}
We performed an experiment to train the domain classifier with and without augmented features when $Z_{c,d}$ is not discarded.

We give each augmented feature $\tilde{Z}$ a soft domain label. Suppose $N$ domains are present in training. 
the original feature $Z$ is from domain $d$. $Z_i$ is from domain $d_i$ and $Z_j$ is from domain $d_j$. The ratio to interpolate $Z_{c,d}$ with $Z_{i_{c,d}}$ is $\lambda_1$, and the ratio to interpolate $Z_{\neg c, d}$ with $Z_{j_{c,d}}$ is $\lambda_2$. 
The domain label of $\tilde{Z}$, $\tilde{d}\in \mathbb{R}^{N}$ at position $d$ is $\frac{\lambda_1+\lambda_2}{2}$. The value of $\tilde{d}$ at position $d_i$ is $\frac{1-\lambda_1}{2}$, and at position $d_j$ is $\frac{1-\lambda_2}{2}$.
Other positions are set to 0. Binary cross-entropy loss is used in the training.

Table \ref{tab:domain_augmented} shows the results on the Camelyon17 dataset. Classification accuracy on the test domain is reported. It suggests that training the domain classifier with augmented features could harm the domain generalization performance. Augmented features may not follow the distribution of existing domains.  

\begin{table}[!ht]
    \centering
    \begin{tabular}{l|c}
    \hline
          Training domain classifier & Test domain accuracy \\ \hline
          with aug features & 77.4$\pm$7.1 \\
          without aug features & \textbf{79.6$\pm$7.0}\\\hline
    \end{tabular}
    \caption{Training domain classifier with and without augmented features on the Camelyon17 dataset.}
    \label{tab:domain_augmented}
\end{table}

\section{Additional Results on Domain Generalization Performance}
We include the domain generalization performance of XDomainMix on two more datasets, VLCS and OfficeHome. 
\begin{itemize}
   \item VLCS \cite{vlcsfang2013unbiased}. This dataset contains 10,729 photos of 5 classes from 4 existing datasets (domains). 
    
   \item OfficeHome \cite{venkateswara2017deep}. This dataset contains 15,588 images of 65 office and home objects in 4 visual styles (domains): art painting, clipart, product (images without background), and real-world (images captured with a camera). 
\end{itemize}

In addition, we include the results of two methods that were proposed earlier. 
\begin{itemize}
    \item CORAL \cite{sun2016deep} aligns the second-order statistics of the representations across different domains.
    \item IRM \cite{arjovsky2019invariant} learns a representation such that the optimal classifier matches all domains.
\end{itemize}

To perform experiments on VLCS and OfficeHome, We follow the setup in DomainBed, and use a pre-trained ResNet-50. Each domain in the dataset is used as a test domain in turn, with the remaining domains serving as training domains.  
Hyperparameters are the same as what we used for PACS and TerraIncognita. 
The best model is selected based on its performance on the validations split of the training domains.
The averaged classification accuracy of the test domains over 3 runs is reported. 

Table \ref{tb:result_vlcs} shows the result. 
Our method does not perform well on VLCS and ranks third on the OfficeHome dataset. Overall, our method still yields the highest average performance on benchmark datasets. 

\begin{table*}[ht!]
    \centering
    \begin{tabular}{l|lllllll|c}
        \hline
        Method  & Camelyon17      & FMoW            & PACS            & VLCS           & OfficeHome            & TerraIncognita  & DomainNet     & Average \\ \hline
        ERM     & 70.3$\pm$6.4    & 32.3$\pm$1.3    & 85.5$\pm$0.2    & 77.5$\pm$0.4   & 66.5$\pm$0.3          & 46.1$\pm$1.8    & 43.8$\pm$0.1  & 60.3        \\
        CORAL   & 59.5$\pm$7.7    & 32.8$\pm$0.7    & 86.2$\pm$0.3    & \textbf{78.8$\pm$0.6}   & \textbf{68.7$\pm$0.3} & 47.6$\pm$1.0    & 41.5$\pm$0.1  & 59.3       \\
        IRM     & 64.2$\pm$8.1    & 30.0$\pm$1.4    & 83.5$\pm$0.2    & 78.5$\pm$0.5   & 64.3$\pm$2.2          & 47.6$\pm$0.8    & 33.9$\pm$2.8  & 57.4       \\
        GroupDRO& 68.4$\pm$7.3    & 30.8$\pm$0.8    & 84.4$\pm$0.8    & 76.7$\pm$0.6   & 66.0$\pm$0.7          & 43.2$\pm$1.1    & 33.3$\pm$0.2  & 57.5        \\
  
        RSC     & 77.0$\pm$4.9\^  & 32.6$\pm$0.5\^  & 85.2$\pm$0.9    & 77.1$\pm$0.5   & 65.5$\pm$0.9          & 46.6$\pm$1.0    & 38.9$\pm$0.5  & 60.4       \\
        MixStyle& 62.6$\pm$6.3\^  & 32.9$\pm$0.5\^  & 85.2$\pm$0.3    & 77.9$\pm$0.5   & 60.4$\pm$0.3          & 44.0$\pm$0.7    & 34.0$\pm$0.1 & 56.7        \\
        DSU     & 69.6$\pm$6.3\^  & 32.5$\pm$0.6\^  & 85.5$\pm$0.6\^  & 77.2$\pm$0.5\^ & 65.7$\pm$0.4\^        & 41.5$\pm$0.9\^  & 42.6$\pm$0.2\^ & 59.2     \\
        LISA    & 77.1$\pm$6.5    & 35.5$\pm$0.7    & 83.1$\pm$0.2\^  & 76.8$\pm$1.0\^ & 67.4$\pm$0.2\^        & 47.2$\pm$1.1\^  & 42.3$\pm$0.3\^ & 61.3      \\
        Fish    & 74.7$\pm$7.1    & 34.6$\pm$0.2    & 85.5$\pm$0.3    & 77.8$\pm$0.3   & 68.6$\pm$0.4          & 45.1$\pm$1.3    & 42.7$\pm$0.2  & 61.3       \\ 
        \textbf{XDomainMix} & \textbf{80.9$\pm$3.2} & \textbf{35.9$\pm$0.8} & \textbf{86.4$\pm$0.4}  & 76.3$\pm$0.5   & 68.1$\pm$0.2 & \textbf{48.2$\pm$1.3}    & \textbf{44.0$\pm$0.2} & \textbf{62.8}    \\ \hline
    \end{tabular}
   \caption{Performance of XDomainMix compared with state-of-the-art methods performance.  Results with \^{} are produced by us. }
    \label{tb:result_vlcs}
\end{table*} 

\section{Domain Generalization Experiment Details}
Details of the domain generalization experiments of state-of-the-art methods produced by us are as below.
\paragraph{Camelyon17 and FMoW dataset} We run the experiments of RSC, MixStyle, and DSU in the testbed provided by LISA.
Following the instruction of the publisher of the datasets, non-pretrained DenseNet-121 is used as the backbone for Camelyon17. Pretrained DenseNet-121 is used as the backbone for FMoW.

Our implementation of RSC follows that in DomainBed. 
The drop factors are set to 1/3, the value recommended in RSC.

 For MixStyle, our implementation is based on the code published by the author. 
 MixStyle module is inserted after the first and second DenseBlock. 
 All hyperparameters are set to be the value recommended by the author. 
 The MixStyle mode is random, which means that feature statistics are mixed from two randomly drawn features. 
 The probability of using MixStyle is set to 0.5. 
$\alpha$, the parameter of the Beta distribution is set to 0.1.
 The scaling parameter to avoid numerical issues, $\epsilon$ is set to 1e-6.

We implement DSU following the published code by the author. All hyperparameters are set to be the value recommended by the author.
DSU module is inserted after the first convolutional layer, first maxpool layer, and every transition block. 
The probability of using DSU is set to 0.5.
The scaling parameter to avoid numerical issues, $\epsilon$ is set to 1e-6.

For the three experiments, the batch size is set to 32, and the model is trained for the same number of epochs as we used to train our method. 
We tune the learning rate in \{1e-4, 1e-3, 1e-2\} and select the optimal learning rate based on the performance on the validation domain.  
Weight decay is set to 0.

\paragraph{PACS, VLCS, OfficeHome, TerraIncognita and DomainNet dataset}
We use DomainBed as the testbed to experiment DSU and LISA. ResNet-50 pretrained on ImageNet is used as the backbone.

Our implementation of LISA is based on the code published by the author. Following the author's suggestion, CuMix is used to mix up input samples. The mix\_up alpha is set to 2.

The official code of DSU for ResNet-50 is used. Other settings are the same as what we used for Camelyon17 and FMoW experiments.

The model is trained for the same number of steps as we used to train our method. 
We tune the learning rate, weight decay, and batch size in the range listed by DomainBed. The learning rate is tuned in (1e-5, 1e-3.5). The weight decay is tuned in (1e-6, 1e-2). The batch size is tuned in $[32, 45]$  (24 for DomainNet). 20 groups of hyperparameters are searched. The optimal hyperparameters are selected based on the performance of the validation split of training domains. 

\end{document}